\documentclass[11pt]{article}
\usepackage{amsmath}
\usepackage{amsfonts}
\usepackage{placeins} 

\usepackage[preprint]{acl}

\usepackage{times}
\usepackage{latexsym}

\usepackage[T1]{fontenc}

\usepackage[utf8]{inputenc}

\usepackage{microtype}

\usepackage{inconsolata}

\usepackage{graphicx}
\usepackage{booktabs}
\usepackage{subcaption}
\usepackage{hyperref}
\usepackage{xurl}

\title{Sparse Transformer: a Controlled  Testbed \\ for Mechanistic Interpretability}

\author{Anton Korznikov\thanks{Correspondence: \texttt{korznikovantona@gmail.com}} \quad Alexey Dontsov \quad Elena Tutubalina \\[1ex]
  \parbox{\textwidth}{\centering
    Code: \href{https://github.com/AntonKorznikov/SparseTransformer}{github.com/AntonKorznikov/SparseTransformer}\\[0.3ex]
    Models: \href{https://huggingface.co/AntonKorznikov/SparseTransformer}{huggingface.co/AntonKorznikov/SparseTransformer}}
}

\begin{document}
\maketitle

\begin{abstract}

Mechanistic interpretability aims to decompose neural network computations into human-understandable features, yet progress is limited by the absence of ground-truth units to evaluate against. We address this with a Sparse Transformer (ST): a 5-layer language model trained on TinyStories whose wide, TopK-sparse MLP forces neuron specialization. We show that ST neurons are highly interpretable: they outperform a loss-matched dense transformer by 43\% on AutoInterp, and circuits of a median of 12 neurons suffice to reproduce individual model predictions. This testbed allows us to  directly compare feature learning methods: Transcoder, which approximates the MLP input-output mapping, recovers ST's neuron vectors with 52.6\% higher cosine similarity than Sparse Autoencoder, which only reconstructs outputs. Furthermore, the ST enables a controlled study of polysemanticity: distilling its sparse MLP into a dense counterpart reveals how compression leads to feature superposition. These results show how a known reference basis sharpens evaluation and understanding, a role we hope the ST can serve for the field.
\end{abstract}

\section{Introduction}
\label{sec:intro}

\begin{figure*}[t!]
  \centering
  \includegraphics[width=\linewidth]{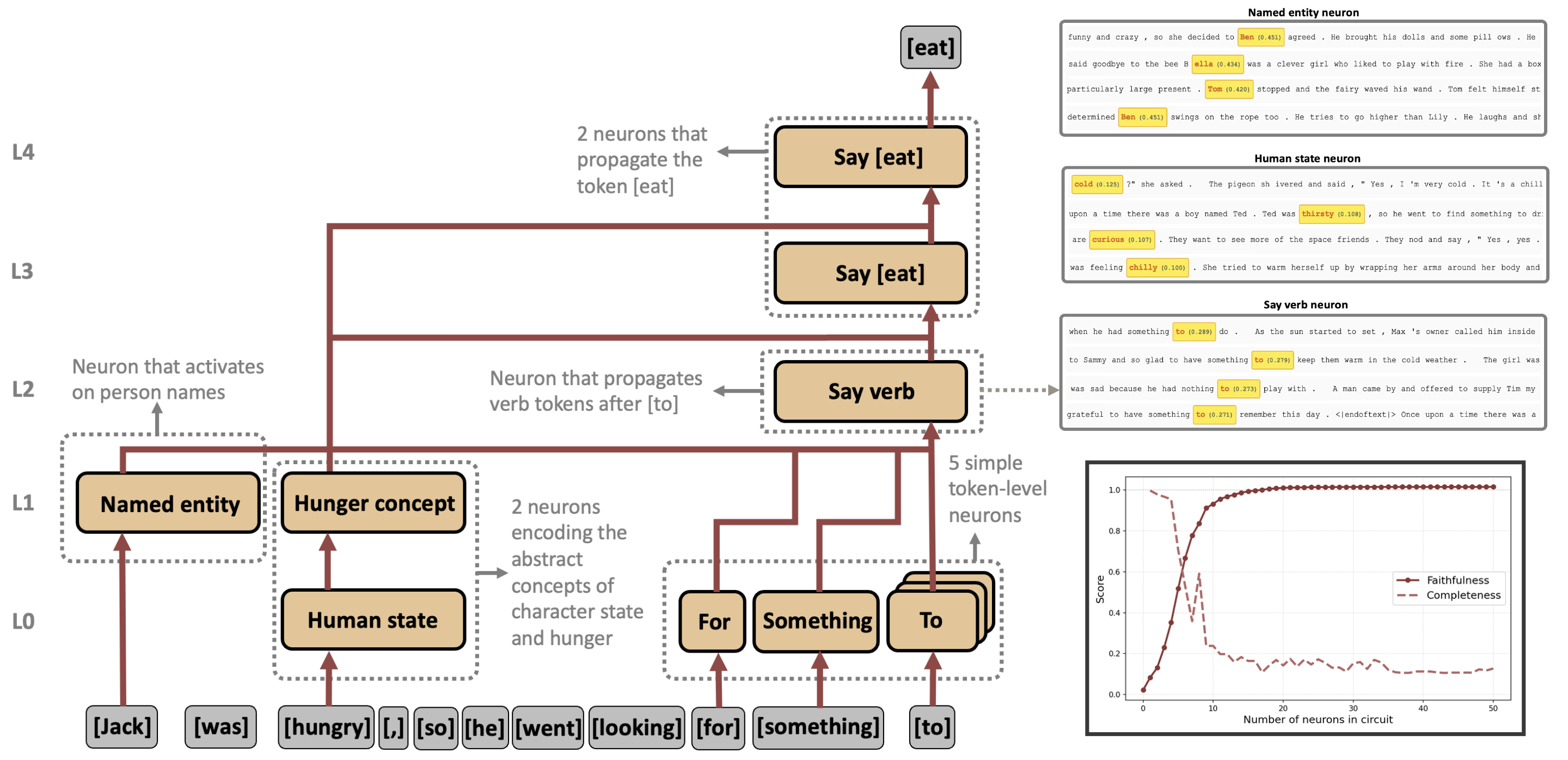}
  \caption{\textbf{An interpretable 11-neuron circuit.} Given the prompt \emph{``Jack was hungry so he went looking for something to''}, nodes are MLP neurons grouped by layer (L0--L4); edges indicate causal influence. \textbf{(top right)} Neuron dashboards show top activating contexts for 3 circuit nodes. \textbf{(bottom right)} Faithfulness and completeness as neurons are added by beam search; 11 neurons suffice to reach faithfulness $\geq 0.95$.}
  \vspace{-1em}
  \label{fig:circuit}
\end{figure*}

Language model neurons are \textit{polysemantic}: each responds to many unrelated inputs, making them resistant to human interpretation \citep{elhage2022toy, olah2020zoom}. Mechanistic interpretability aims to address this, yet the field faces a persistent methodological bottleneck: the absence of a reliable ground-truth \citep{sharkey2025open, chanin2026synthsaebench}. When Sparse Autoencoder (SAE) \citep{bricken2023monosemanticity} decomposes a model's residual stream into a sparse sum of monosemantic features, we cannot know whether those features correspond to genuine computational units or are artifacts of the decomposition method \citep{korznikov2026sanity, leask2025sparse}. Similarly, when a circuit is identified as responsible for a behavior, faithfulness metrics remain proxies, as the true circuit is never known \citep{sutter2026non}.
 
Existing attempts to obtain ground truth rely on toy models with hand-crafted 
tasks \citep{elhage2022toy, nanda2023progress, huben9research, sharkey2022taking}, 
which do not process natural language, leaving it unclear how findings transfer 
to real models. A more principled approach is to train a model whose internal structure is 
interpretable \emph{by design}, so that its known units can serve as a reference 
for interpretability methods on natural language.

We propose the Sparse Transformer (ST) as such a model. ST is a decoder-only 
transformer whose MLP layers apply a TopK gate after the ReLU nonlinearity: at 
each forward pass, only $K$ out of $N = M \cdot d_\text{model}$ neurons activate 
($M$ is the width multiplier). When $M$ is large and $K$ is small, each neuron 
must specialise, with sparsity acting as an inductive bias towards 
monosemanticity. Thus, ST provides an operational ground truth in two 
complementary senses. The first is structural: because exactly $K$ neurons fire 
per pass, the MLP output is a sum of $K$ known value vectors from a known active 
set, matching the sparse decomposition that SAEs and Transcoders are designed to 
recover. The second is semantic: we verify empirically that these neurons are 
monosemantic and human-interpretable. Together, these properties 
make the ST a controlled testbed for evaluating feature-recovery methods on 
natural language.
 
We make the following contributions:
 
\begin{itemize}

  \item We show that our primary ST model is \textbf{highly interpretable}: it scores $0.866$ on AutoInterp (vs.\ $0.605$ for a loss-matched dense transformer) and admits compact circuits that reproduce individual model predictions.

  \item In \textbf{Case Study \#1}, we use the ST's known neuron basis to compare SAEs and Transcoders, finding that Transcoders reconstruct MLP outputs 3.8$\times$ more accurately (NMSE) and recover true neurons with 
  52.6\% higher cos-sim.

  \item In \textbf{Case Study \#2}, we distil the ST's MLP into a dense counterpart and study polysemanticity in a controlled setting, finding each dense neuron superposes $\sim$150 sparse neurons with its weight vectors reflecting this mixture.

  \item Additionally, we report an empirical \textbf{width–sparsity trade-off} suggesting that strong sparsity can be offset by moderate width expansion, showing a path toward larger interpretable LLMs.

\end{itemize}

Together, these contributions establish the ST as a testbed that bridges the gap between toy models and production LLMs: trained on natural language, yet with known computational units at every layer. We publicly release all 35 models trained across the width–sparsity grid, along with checkpoints at every 10\% of training, to support systematic study of feature formation, circuit emergence, and interpretability methods evaluation.

\section{Related Work}
\label{sec:related}

\paragraph{Mechanistic Interpretability.}
Mechanistic interpretability aims to reverse-engineer the computations inside neural networks into human‑understandable algorithms \citep{olah2020zoom, elhage2021mathematical}. Its core primitives are \textit{features}, interpretable properties of the input that a model represents (e.g., ``past tense'', ``character names'', ``sentiment''), and \textit{circuits}, groups of features that interact to produce a specific behavior \citep{wang2022interpretability, tigges2024llm, lindsey2025biology, marks2025sparse}. The most natural feature unit would be a single neuron, but language-model neurons are typically \textit{polysemantic}, responding to many unrelated concepts, which motivates the search for more disentangled representations \cite{olah2017feature, bricken2023monosemanticity, kopf2026capturing}.

\paragraph{SAEs and Transcoders.}

The polysemanticity has motivated methods that decompose dense activations into sparse, interpretable features. These methods rest on the \emph{superposition hypothesis}~\citep{elhage2022toy, olah2020zoom}: any dense activation vector $\mathbf{x} \in \mathbb{R}^{d_{\text{model}}}$ can be uniquely represented as a sparse linear combination of feature vectors. Both Sparse Autoencoders (SAEs)~\citep{bricken2023monosemanticity, rajamanoharan2024jumping, lieberum2024gemma, gao2025scaling} and Transcoders~\citep{dunefsky2024transcoders, ameisen2025circuit, paulo2025transcoders} operationalize this by learning a dictionary whose size $D$ is much larger than the model's hidden dimension ($D \gg d_{\text{model}}$), while enforcing that only a small number $K \ll d_{\text{model}} $ of features are active on any input. This sparsity pressure forces each latent to specialize to a coherent, human-understandable concept~\citep{cunningham2023sparse}.

An SAE reconstructs a target dense activation vector $\mathbf{x}$ (e.g., from the residual stream or an MLP output) via $\hat{\mathbf{x}} = \mathbf{W}^{\text{dec}} \text{TopK}\bigl(\mathbf{W}^{\text{enc}} \mathbf{x} + \mathbf{b}^{\text{enc}}\bigr) + \mathbf{b}^{\text{dec}}$, where $\mathbf{W}^{\text{enc}} \in \mathbb{R}^{D \times d_{\text{model}}}$ and $\mathbf{W}^{\text{dec}} \in \mathbb{R}^{d_{\text{model}} \times D}$. The TopK gate retains only the $K$ largest latent coefficients and zeros the rest, so the reconstruction is a sparse sum of $K$ selected columns from $\mathbf{W}^{\text{dec}}$.

A Transcoder usually adopts the same TopK-sparse, overcomplete architecture but targets the functional behavior of a component (e.g. MLP) rather than its output representation. Given a module's input $\mathbf{x}$, it approximates the module's output $\mathbf{y}$ as $\hat{\mathbf{y}} = \mathbf{W}_{\text{out}} \text{TopK}\bigl(\mathbf{W}_{\text{in}} \mathbf{x}_{\text{in}}+\mathbf{b}_{\text{in}}\bigr) + \mathbf{b}_{\text{out}}$. Because the Transcoder must reproduce the full input-output mapping rather than merely reconstruct activations, it is conjectured to learn more mechanistically faithful latents~\citep{paulo2025transcoders}. However, both methods only claim to recover the true sparse decomposition, and without ground truth one cannot verify whether discovered latents are genuine features or geometric artifacts of reconstruction ~\citep{korznikov2026sanity, leask2025sparse}. In Section~\ref{sec:cs1}, we use the Sparse Transformer's known neuron basis to directly test which method more faithfully recovers the ground-truth MLP vectors.

\paragraph{Evaluating Interpretability Methods.}
Rigorous evaluation of feature extraction and circuit discovery requires known ground-truth units. Two categories of proxies exist: synthetic activation settings where sparse features are explicitly defined \citep{sharkey2022taking, chanin2025sparse, korznikov2026sanity, chanin2026synthsaebench}, and toy models with hand-crafted tasks such as modular arithmetic \citep{nanda2023progress}, board-state prediction \citep{huben9research}, or simple superposition \citep{elhage2022toy}. Both categories share a key limitation: they do not process natural language, making it unclear how findings transfer to real models. ST bridges this gap with a known reference basis in a language model.


\paragraph{Sparse Architectures and Interpretability.}

\citet{elhage2022solu} introduce SoLU to reduce average neuron activity,
but soft sparsity provides no monosemanticity guarantee: as the authors note,
reduced activations can mask polysemantic neurons rather than eliminate them.
\citet{gao2025weight} enforce weight sparsity to yield compact, inspectable circuits, a different inductive bias than activation sparsity.
Mixture-of-Experts models \citep{shazeer2017outrageously, fedus2022switch} and
sparse attention \citep{child2019generating, zhao2019explicit} reduce FLOPs by
activating subsets of parameters or positions, but do not target interpretability
and their activated units remain polysemantic \citep{guo2026advancing}. ST differs from all of these: a TopK gate enforces \emph{exact} activation sparsity within the MLP, keeping precisely $K$ neurons active per forward pass to force specialisation for interpretability.
 
\section{Sparse Transformer}
\label{sec:st}

\subsection{Architecture}
\label{sec:arch}

Our ST is a decoder-only transformer \citep{vaswani2017attention} with 5 layers, hidden size $d_{\text{model}}{=}256$, and 8 attention heads. We use RoPE positional embeddings \citep{su2024roformer}, untied input/output embeddings, and RMSNorm \citep{zhang2019root}. The MLP at each layer has $N = M \cdot d_{\text{model}}$ neurons ($M$ is the width multiplier) and computes
\[
  \text{MLP}(\mathbf{x}) = V_{\text{out}} \cdot \text{TopK}\bigl(\text{ReLU}(V_{\text{in}} \mathbf{x})\bigr),
  \label{eq:mlp}
\]
where $\text{TopK}(\cdot)$ retains the $K$ largest activations and zeros the rest, following the same mechanism and motivation as TopK SAEs \citep{gao2025scaling}. The output is thus exactly a sparse sum of $K$ value vectors $\mathbf{v}_i{=}V_{\text{out}}[:,i]$.

\begin{figure*}[t!]
  \centering
  \begin{subfigure}[b]{0.33\textwidth}
    \includegraphics[width=\linewidth]{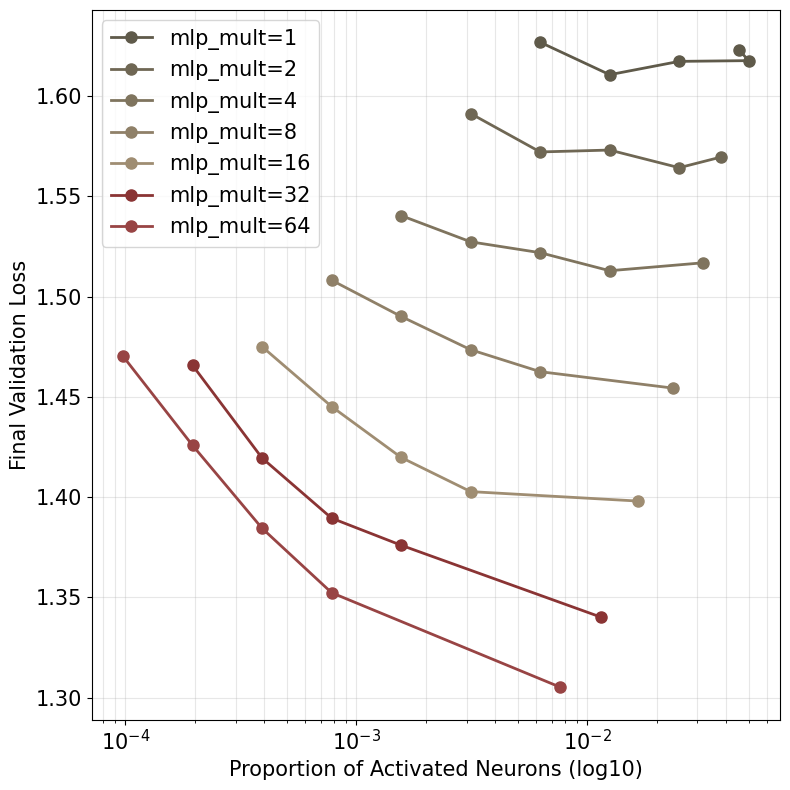}
    \label{fig:prop_activated}
  \end{subfigure}\hfill
  \begin{subfigure}[b]{0.33\textwidth}
    \includegraphics[width=\linewidth]{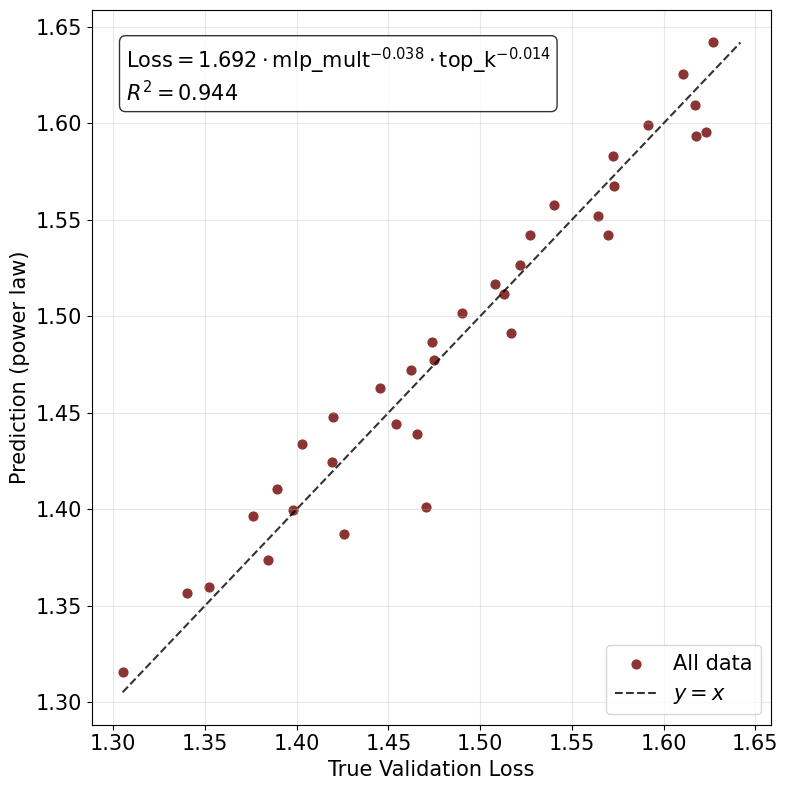}
    \label{fig:powerlaw}
  \end{subfigure}\hfill
  \begin{subfigure}[b]{0.33\textwidth}
    \includegraphics[width=\linewidth]{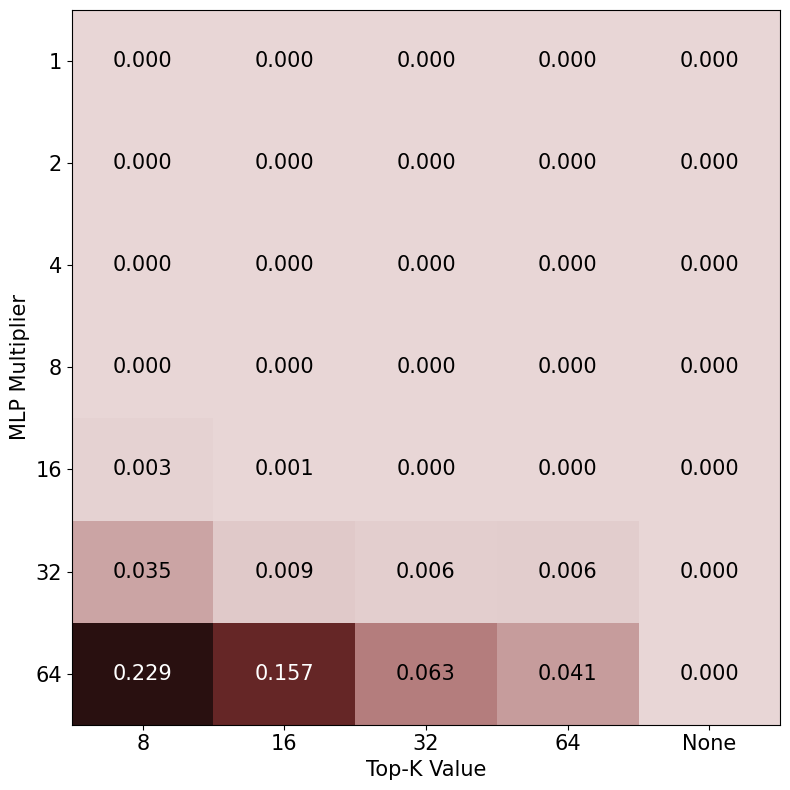}
    \label{fig:heatmap_dead}
  \end{subfigure}
  \vspace{-1em}
  \caption{\textbf{Width–sparsity trade-off.} \textbf{(left)} Validation loss vs.\ proportion of activated neurons for each MLP width multiplier $M$; loss decreases monotonically with both width and activation rate. \textbf{(middle)} Power-law predictions vs.\ true validation loss across all 35 models ($R^2{=}0.944$). \textbf{(right)} Fraction of dead neurons across the $(M,K)$ grid; negligible for $M \leq 16$ but rising at large width and small $K$.}
  \label{fig:scaling}
  \vspace{-1em}
\end{figure*}

\subsection{Training Setup}
\label{sec:training}

We train on the TinyStories dataset \citep{eldan2023tinystories}, a corpus of short children's stories generated by GPT-4 \citep{achiam2023gpt} that is rich enough for meaningful language modelling while being tractable for a small model. 

We train the full Cartesian product of:
\[
\begin{aligned}
  M &\in \{1,2,4,8,16,32,64\} \\
  K &\in \{8,16,32,64,\text{None}\}
\end{aligned}
\]
yielding 35 models. $K{=}\text{None}$ removes the TopK constraint entirely, providing a dense baseline.

All models are trained for $300{,}000$ steps (${\approx}4.9$B tokens) with AdamW \citep{loshchilov2017decoupled}, batch size 64, and a cosine schedule decaying learning rate from $3\times10^{-4}$ to $10^{-5}$. Context length is 256 tokens. To prevent dead neurons, we apply neuron resampling every $1{,}000$ steps following standard SAE practice \citep{bricken2023monosemanticity, gao2025scaling}: any neuron with activation frequency below $5\times10^{-6}$ is randomly reinitialised. All 35 models share the same random seed and data order. We release all checkpoints, including snapshots at every $10\%$ of training.

\subsection{Width–Sparsity Trade-off}
\label{sec:scaling}

We first characterize how MLP width multiplier ($M$) and activation sparsity ($K$) jointly determine language modeling performance. We find that the $35$-entry validation loss matrix across the $(M,K)$ grid is nearly rank-1 (first singular values: $\sigma_1{=}7.91$, $\sigma_2{=}0.08$), indicating that $M$ and $K$ contribute as approximately separable, multiplicative factors. This motivates a power-law fit for the final validation loss:
$$
  \mathcal{L}(M,K) = c \cdot M^{\alpha} \cdot K^{\beta},
$$
following the functional form commonly used in LLMs scaling \citep{kaplan2020scaling, hoffmann2022training}.
The model achieves $R^2{=}0.944$ (Fig.~\ref{fig:scaling}, middle), substantially outperforming linear and logarithmic alternatives, indicating strong predictive power. Fitted coefficients are shown in Table~\ref{tab:powerlaw}.

\begin{table}[t!]
  \centering
  \begin{tabular}{lcc}
    \toprule
    \textbf{Parameter} & \textbf{Estimate} & \textbf{95\% CI} \\
    \midrule
    $c$       & $1.696$ & $[1.667,\ 1.725]$ \\
    $\alpha$  & $-0.038$ & $[-0.041,\ -0.034]$ \\
    $\beta$   & $-0.015$ & $[-0.018,\ -0.013]$ \\
    \bottomrule
  \end{tabular}
  \caption{Power-law fit coefficients for $\mathcal{L}(M,K)$.}
  \vspace{-0.5em}
  \label{tab:powerlaw}
\end{table}

\paragraph{Interpretation.} 
The exponents $\alpha{=}{-}0.038$ and $\beta{=}{-}0.015$ show that validation loss is more sensitive to MLP width than to sparsity: the ratio $\beta/\alpha \approx 0.4$ means width must grow as $K^{0.4}$ to hold loss constant. Concretely, within our grid, tightening sparsity from $K{=}64$ to $K{=}8$ requires only a $\sim$2.3$\times$ increase in $M$ (since $8^{0.4}\approx 2.3$). This sub-linear trade-off appears achievable in practice: Mixture-of-Experts architectures \citep{mu2025comprehensive} routinely expand effective MLP width by $8$--$64\times$ without proportionally increasing compute, suggesting that highly sparse, and thus interpretable, transformers may not need to sacrifice language modelling performance.

\paragraph{Dead neurons and model selection.} Following standard SAE practice \citep{bricken2023monosemanticity}, a neuron is counted as \emph{dead} if its activation frequency falls below $10^{-6}$ over the validation set. Dead neurons are negligible for $M \leq 32$ but rise to $22.9\%$ at $(M{=}64, K{=}8)$ and $15.7\%$ at $(M{=}64, K{=}16)$ (Figure~\ref{fig:scaling}, right). We select $M{=}32$, $K{=}8$ for all subsequent experiments: the sparsest configuration with an acceptably low dead-neuron rate ($3.5\%$), achieving validation loss $1.466$, in the same range as the 5-layer models of \citet{eldan2023tinystories}.

\section{Interpretability of Sparse Transformer}
\label{sec:interp}

Having established the scaling properties of the ST, we now verify that our primary ST model ($M{=}32$, $K{=}8$) achieves its core objective: being highly interpretable at the neuron and circuit level. 

\subsection{Monosemanticity}
\label{sec:mono}

We quantify interpretability with the AutoInterp score, which has been shown to align closely with human assessments of neuron interpretability \citep{paulo2024automatically, karvonen2025saebench}. For each of 100 randomly sampled neurons, GPT-4o-mini \citep{hurst2024gpt} first describes the neuron's top-activating contexts, then uses only that description to classify 20 held-out examples (10 active, sampled proportionally to activation strength, and 10 silent). The score is the mean classification accuracy over the 100 neurons.

As presented in Table~\ref{tab:autointerp} our ST (M=32, K=8) reaches an AutoInterp score of 0.866, providing quantitative evidence that its neurons are substantially more interpretable. By contrast, a dense model that achieves the same validation loss ($M{=}4$, $K{=}\text{None}$) scores only 0.605, while a minimal dense baseline (M=1, K=None) scores 0.565, barely above the 0.5 chance level. Even at matched width ($M{=}32$, $K{=}\text{None}$), the dense model scores only 0.667, so the gain is attributable to sparsity rather than width. A 43\% relative improvement over the loss-matched dense baseline suggests that TopK sparsity yields substantially more interpretable neurons in this setting. 

To complement the AutoInterp score, we manually constructed activation dashboards for 20 randomly sampled neurons. Each dashboard shows top-activating contexts at multiple intensity tiers. Each of the 20 sampled neurons activates on a specific, human-interpretable concept: for example, L0-N7792 responds to planning verbs (\emph{plan, brainstorm, discuss}), L2-N3125 fires at the moral lesson at the end of a story, and L4-N4150 activates on loud sounds (\emph{loud noise, loud bang}). A full set of 20 dashboards is provided in App.~\ref{app:dashboards}, with the complete neuron set available in the released repository.

\begin{table}[t]
  \centering
  \small
  \setlength{\tabcolsep}{3pt}
  \begin{tabular}{lcc}
    \toprule
    \textbf{Model} & \textbf{AutoInterp} & \textbf{95\% CI} \\
    \midrule
    Random guessing     & 0.500 & --- \\
    Dense ($M{=}1$)     & 0.565 & [0.51,\ 0.61] \\
    Dense ($M{=}4$, loss-matched)     & 0.605 & [0.55,\ 0.65] \\
    Dense ($M{=}32$, width-matched)     & 0.667 & [0.61,\ 0.72] \\
    ST ($M{=}32$, $K{=}8$) & \textbf{0.866} & [0.82,\ 0.91] \\
    \bottomrule
  \end{tabular}
  \caption{\textbf{AutoInterp scores.} TopK sparsity yields substantially more interpretable neurons: ST ($M{=}32$, $K{=}8$) scores 0.866 vs. 0.605 for the loss‑matched and 0.667 for the width-matched dense baseline.}
  \vspace{-1em}
  \label{tab:autointerp}
\end{table}

 
\subsection{Circuit Analysis}
\label{sec:circuit}

We ask whether ST computations form complete, human-readable circuits, not just monosemantic neurons in isolation. We focus on an illustrative example from the original TinyStories work \citep{eldan2023tinystories}, the prompt \emph{``Jack was hungry so he went looking for something to''}, where the model assigns probability 0.98 to the next token \texttt{[eat]}.
 
\paragraph{Circuit extraction.} We follow the standard ablation-based protocol \citep{conmy2023towards, arora2026language}: neurons are ranked by their causal effect on $P(\texttt{[eat]})$ and added via beam search to the circuit until two complementary metrics are satisfied. Faithfulness ($F$) measures how well the circuit alone reproduces the full model's behaviour; Completeness ($C$) measures how much performance collapses when the circuit is ablated\footnote{We follow the convention of prior circuit work and define $C$ such that lower is better, with $C=0$ ideal.}:
\[
  F = \frac{P_\text{circuit}(\texttt{[eat]})}{P_\text{full}(\texttt{[eat]})}\qquad
  C = \frac{P_\text{ablated}(\texttt{[eat]})}{P_\text{full}(\texttt{[eat]})}
  \label{eq:fc}
\]
where $P_\text{full}$ is the intact model probability, $P_\text{circuit}$ is the probability when all neurons outside the circuit are zeroed, and $P_\text{ablated}$ is the probability when the circuit neurons themselves are zeroed. A perfect circuit thus has $F{=}1$ and $C{=}0$\footnote{$F$ can exceed $1$ when the remaining ablated neurons are net-suppressive.}.
 
\paragraph{Circuit description.} The resulting circuit is shown in Figure~\ref{fig:circuit} and achieves $F{=}0.95$, $C{=}0.2$ with only 11 neurons. It decomposes into three functional groups: token-level neurons at L0 firing on \texttt{[for]}, \texttt{[something]}, \texttt{[to]}; concept neurons at L0--L1 encoding a character's hunger-related internal state and named-entity identity; and output neurons at L2--L4 propagating verb tokens and directly promoting \texttt{[eat]}. The circuit reads as a human-interpretable program: \emph{a named character in a hunger-related state went looking for something to do $\Rightarrow$ predict a food-consumption verb}. This functional logic is confirmed by targeted ablations within the circuit; for instance, clamping the hunger-concept neuron's activation to $-5\times$ sharply reduced probabilities for \texttt{[eat]}, \texttt{[eating]}, and \texttt{[ate]} while raising those of unrelated verbs like \texttt{[spin]}, \texttt{[point]}, and \texttt{[wonder]}.

This qualitative example illustrates that the sparse neuron basis of our ST can support human-readable, mechanistically grounded circuit analysis in a genuine natural-language model. To confirm this finding is not specific to a single prompt, we extract circuits for 15 additional examples using the same ablation protocol; 14 of them reach $F \geq 0.95$ within 50 neurons, with a median of 12 (App.~\ref{app:circuits}, Tables~\ref{tab:circuit-per-prompt} and~\ref{tab:circuit-grid}). We view this testbed as an invitation to the field to apply systematic circuit discovery methods to natural-language models with a known reference basis.

\begin{figure*}[ht!]
  \centering
  \begin{subfigure}[b]{0.28\linewidth}
    \includegraphics[width=\linewidth]{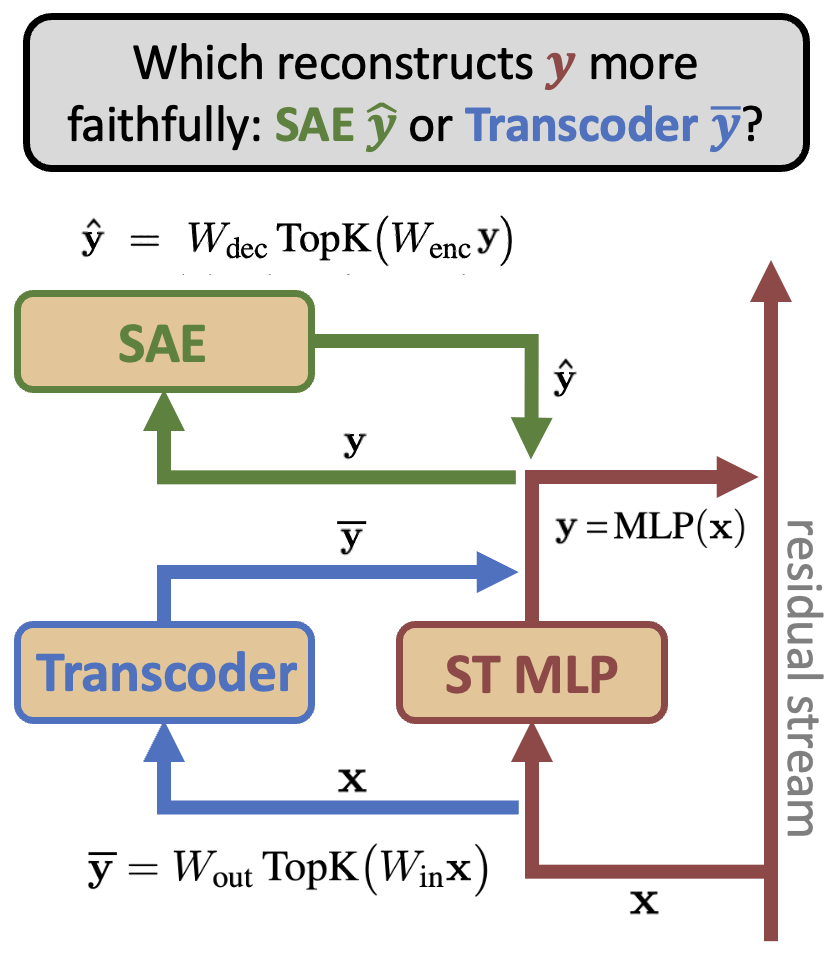}
    \label{fig:nmse_curves}
  \end{subfigure}
  \begin{subfigure}[b]{0.32\linewidth}
    \includegraphics[width=\linewidth]{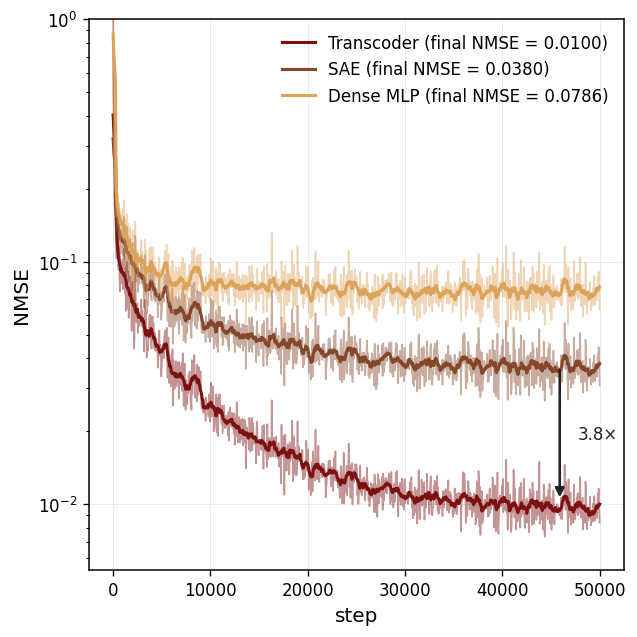}
    \label{fig:nmse_curves}
  \end{subfigure}
  \begin{subfigure}[b]{0.37\linewidth}
    \includegraphics[width=\linewidth]{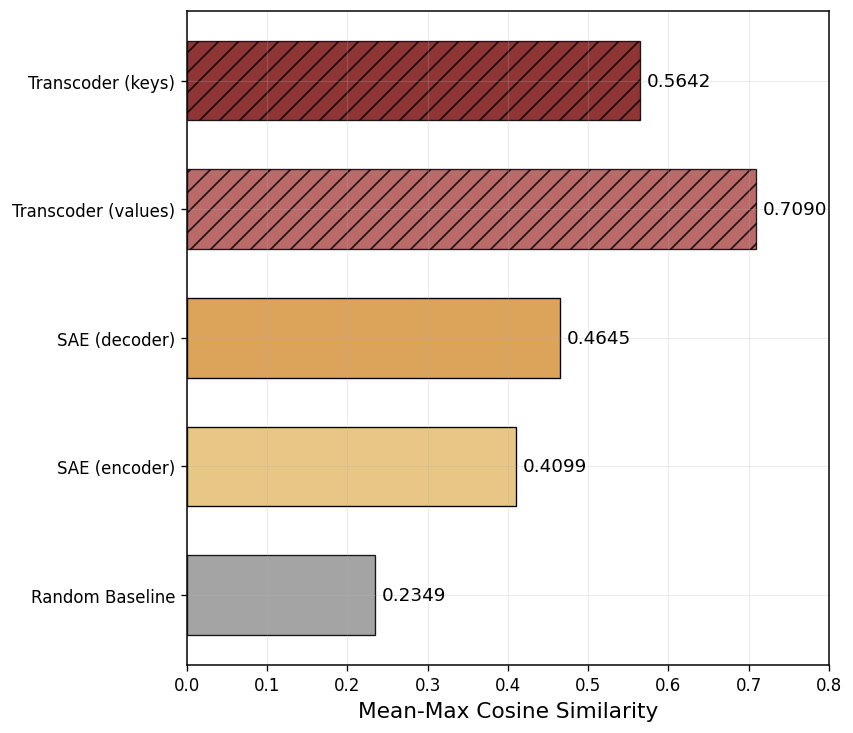}
    \label{fig:cossim_bar}
  \end{subfigure}
 \vspace{-1.5em}
  \caption{\textbf{Feature recovery results (Case Study \#1)}. \textbf{(left)} Schematic of the experiment: comparing Transcoder (trained to approximate the MLP’s input–output mapping) and SAE (trained to reconstruct MLP output activations) against the ground‑truth ST neuron vectors. \textbf{(middle)} NMSE training curves for Transcoder and SAE. Transcoder achieves $3.8\times$ lower final NMSE than SAE. \textbf{(right)} Average cosine similarity between each true ST latent and its closest SAE/Transcoder latent. Transcoder value vectors achieve $+52.6\%$ improvement over SAE decoder vectors.}
  \vspace{-1em}
  \label{fig:cs1}
\end{figure*}
 
\section{Case Study \#1: SAE vs.\ Transcoder}
\label{sec:cs1}

\subsection{Motivation}

A key advantage of the ST is that the true MLP computation is known by construction: the MLP output is a sparse sum of $K$ value vectors $\mathbf{v}_i$, one per active neuron. Recovering exactly these vectors is the stated goal of both SAEs and Transcoders -- an SAE by decomposing the MLP's output activation, a Transcoder by approximating the full input-output mapping. We can therefore measure directly, by cosine similarity to the true $\mathbf{v}_i$, whether each method achieves what it claims.

\subsection{Setup}

We focus on layer 2 of the primary ST ($M{=}32$, $K{=}8$), which captures richer semantic features than early layers (tied to surface statistics) or late layers (tied to next-token prediction) \citep{rogers2020primer, skean2025layer}. Both methods target the output of this MLP layer: the SAE reconstructs it from the output itself, while the Transcoder predicts it from the MLP's own input. They are trained with identical hyperparameters: dictionary size $D{=}8{,}192$ and sparsity $K{=}8$, matching the true MLP's width ($N$) and sparsity ($K$).

For each ST neuron $i$, the ground-truth vectors are its key $\mathbf{k}_i = V_{\text{in}}[i,:]$ (the input pattern that causes it to activate) and its value $\mathbf{v}_i = V_{\text{out}}[:,i]$ (the vector it adds to the residual stream when active). To measure how well a method recovers these vectors, we use mean-max cosine similarity (MMCS) \citep{chanin2025sparse, sharkey2022taking}: for each true vector, we find the closest learned latent, then average over all true vectors. Formally, let $\mathbf{U} = \{\mathbf{u}_i\}_{i=1}^{N}$ be the set of $N$ ground-truth neuron vectors and $\mathbf{R} = \{\mathbf{r}_j\}_{j=1}^{D}$ the set of $D$ learned latents; then: 
\[   \text{MMCS}(\mathbf{U}, \mathbf{R}) = \frac{1}{N} \sum_{i=1}^{N} \max_{j} \cos(\mathbf{u}_i, \mathbf{r}_j). 
\] 
We apply MMCS to match (1) true value vectors $\mathbf{v}_i$ against SAE encoder and decoder vectors, and (2) true key and value vectors $\mathbf{k}_i, \mathbf{v}_i$ against Transcoder input and output weight vectors respectively.

\subsection{Results}

The Transcoder converges to a Normalized Mean Squared Error (NMSE) of $0.01$ compared to $0.038$ for the SAE, a $3.8\times$ improvement (Fig.~\ref{fig:cs1}, middle), and its features align better with ground truth across all comparisons (Fig.~\ref{fig:cs1}, right). For value recovery, Transcoder value vectors achieve an average cosine similarity of $0.709$ with true value vectors, compared to $0.465$ for the SAE decoder (+52.6\%) and $0.410$ for the SAE encoder (+72.9\%). These gaps are consistent and large: the ranking holds for $K{\in}\{4,16\}$ and for BatchTopK \citep{bussmann2024batchtopk} and Matryoshka \citep{bussmann2025learning} variants, each with 3 seeds (App.~\ref{app:cs1}, Table~\ref{tab:cs1robust}).

\begin{figure*}[t!]
  \centering
  \begin{subfigure}[b]{0.55\linewidth}
    \includegraphics[width=\linewidth]{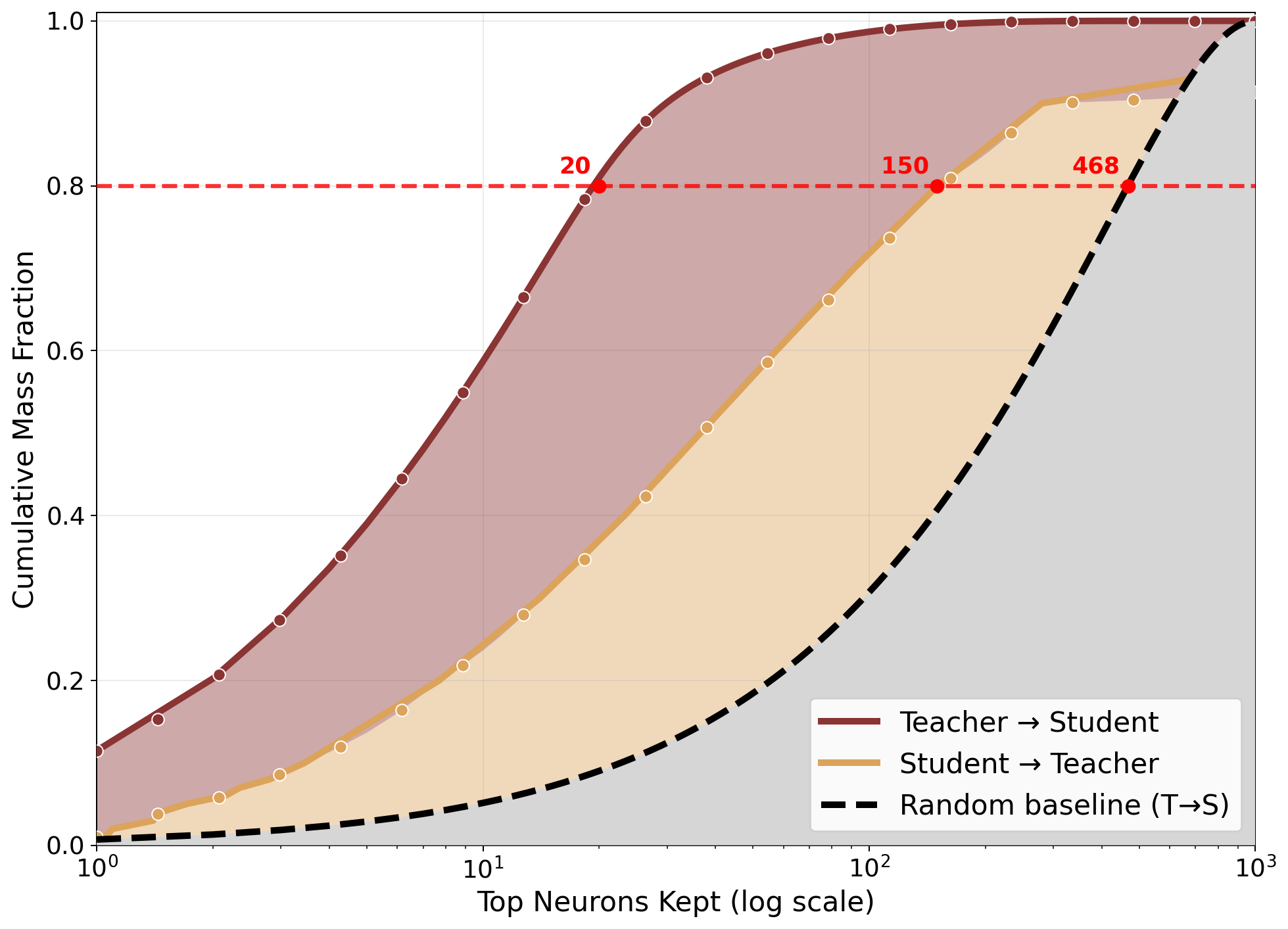}
    \label{fig:cmf}
  \end{subfigure}
  \begin{subfigure}[b]{0.4\linewidth}
    \includegraphics[width=\linewidth]{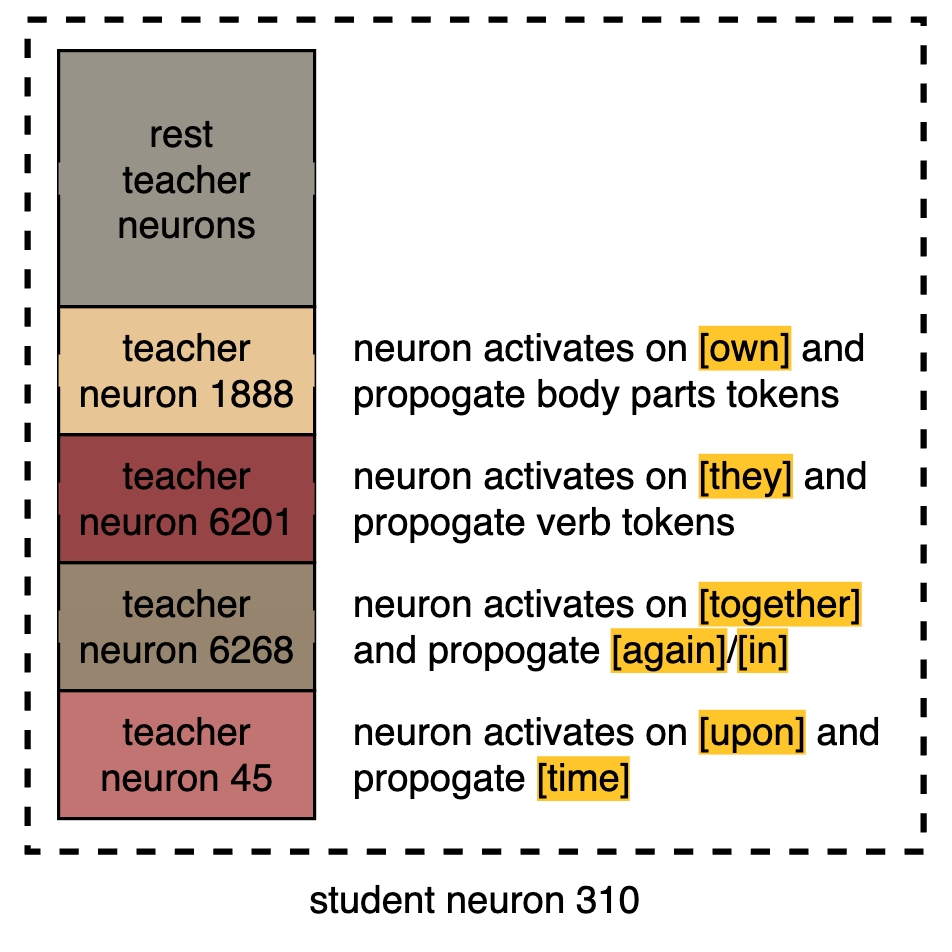}
    \label{fig:polysemantic}
  \end{subfigure}
  \vspace{-1em}
  \caption{\textbf{Teacher-student mapping and polysemanticity.} \textbf{(left)} Cumulative mass functions for both mapping directions. At the $80\%$ level, each teacher neuron $t_i$ maps to ${\sim}20$ student neurons $s_j$, while each student neuron draws from ${\sim}150$ teacher neurons. \textbf{(right)} Student neuron $s_{310}$ superposes four semantically unrelated teacher neurons: body-part tokens, verb propagation after \texttt{[they]}, \texttt{[again]}/\texttt{[in]} propagation, and time-expression tokens.}
  \vspace{-1em}
  \label{fig:cs2_mapping}
\end{figure*}

\subsection{Discussion}

This result may appear counter-intuitive. From an optimisation standpoint, the Transcoder faces the harder task: the SAE receives its target as input, whereas the Transcoder must learn a nonlinear mapping between two spaces, jointly discovering keys and values (see detailed discussion in App.~\ref{app:cs1}). Yet the Transcoder outperforms the SAE on both reconstruction and feature recovery. We attribute this to the stronger inductive bias of the function-level objective: the SAE can achieve low reconstruction error with features that are geometrically convenient but mechanistically unfaithful \citep{korznikov2026sanity}, whereas the Transcoder, constrained to reproduce the MLP's behaviour across the input distribution, cannot resort to such solutions. These findings suggest that function-level methods, such as Transcoders or Crosscoders \citep{ameisen2025circuit}, may recover model features more reliably.

\section{Case Study \#2: Emergence
of Polysemanticity Under Compression}
\label{sec:cs2}
 
\subsection{Motivation}

The superposition hypothesis \citep{elhage2022toy} proposes that polysemanticity arises because a dense MLP approximates the function of a much wider sparse one: each dense neuron superposes multiple sparse features to compensate for reduced width. Prior tests of this account have used toy models where sparse features and weights are defined by the experimenter \citep{elhage2022toy, sharkey2022taking}, a controlled but artificial setting. Our ST provides a more realistic test: sparse features that emerge from training on natural language, with distillation into a dense student serving as a controlled analogue of the compression that produces polysemanticity in production LLMs.

 
\subsection{Setup}

We work with layer 2 of the primary ST ($M{=}32$, $K{=}8$). This layer's sparse MLP (the teacher) has $N_T{=}8{,}192$ neurons $\{t_i\}_{i=1}^{N_T}$. We train a dense student MLP to approximate the teacher's input--output mapping. The student has $N_S{=}1{,}024$ neurons ($8\times$ fewer than the teacher) with no TopK constraint, trained for $50{,}000$ steps on input--output pairs drawn from the teacher. The student is trained solely to minimise MSE on the teacher's input–output pairs, with no language-modelling objective. It achieves NMSE=$0.078$, indicating a faithful approximation of the teacher (Fig.~\ref{fig:cs1}).


\begin{figure*}[ht!]
  \centering
  \begin{subfigure}[b]{0.32\linewidth}
    \includegraphics[width=\linewidth]{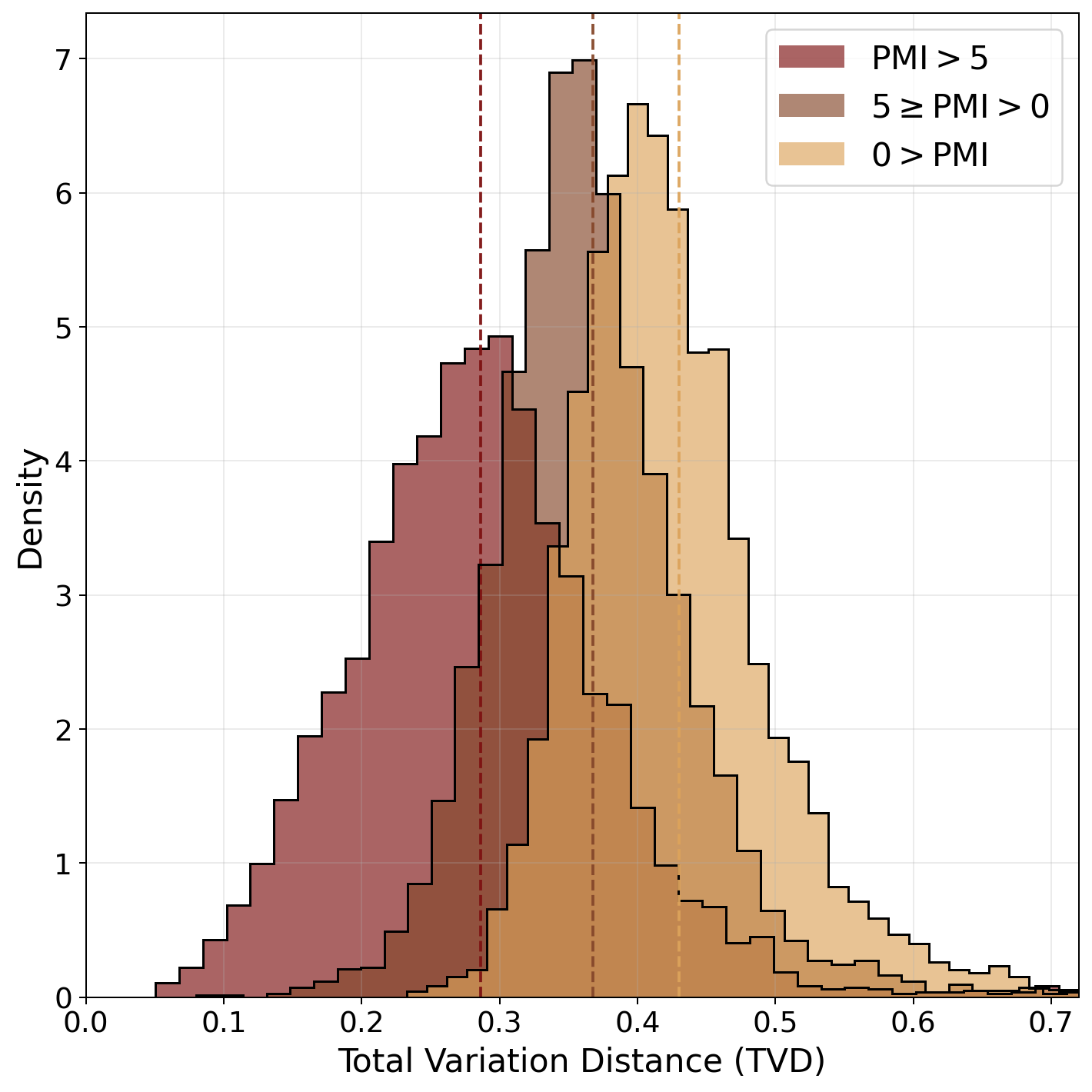}
    \label{fig:tvd}
  \end{subfigure}
  \begin{subfigure}[b]{0.32\linewidth}
    \includegraphics[width=\linewidth]{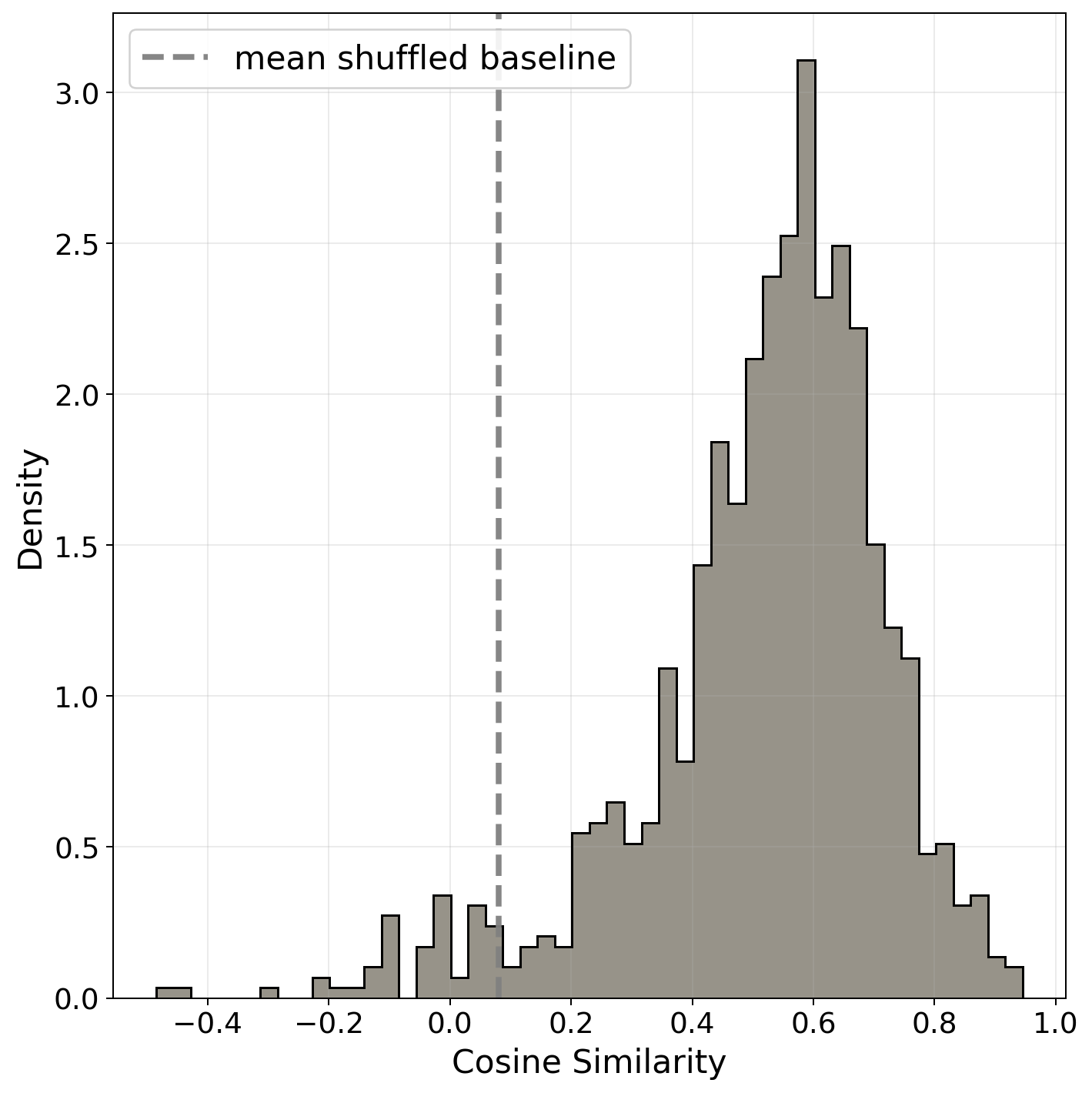}
    \label{fig:geometric_values}
  \end{subfigure}
  \begin{subfigure}[b]{0.32\linewidth}
    \includegraphics[width=\linewidth]{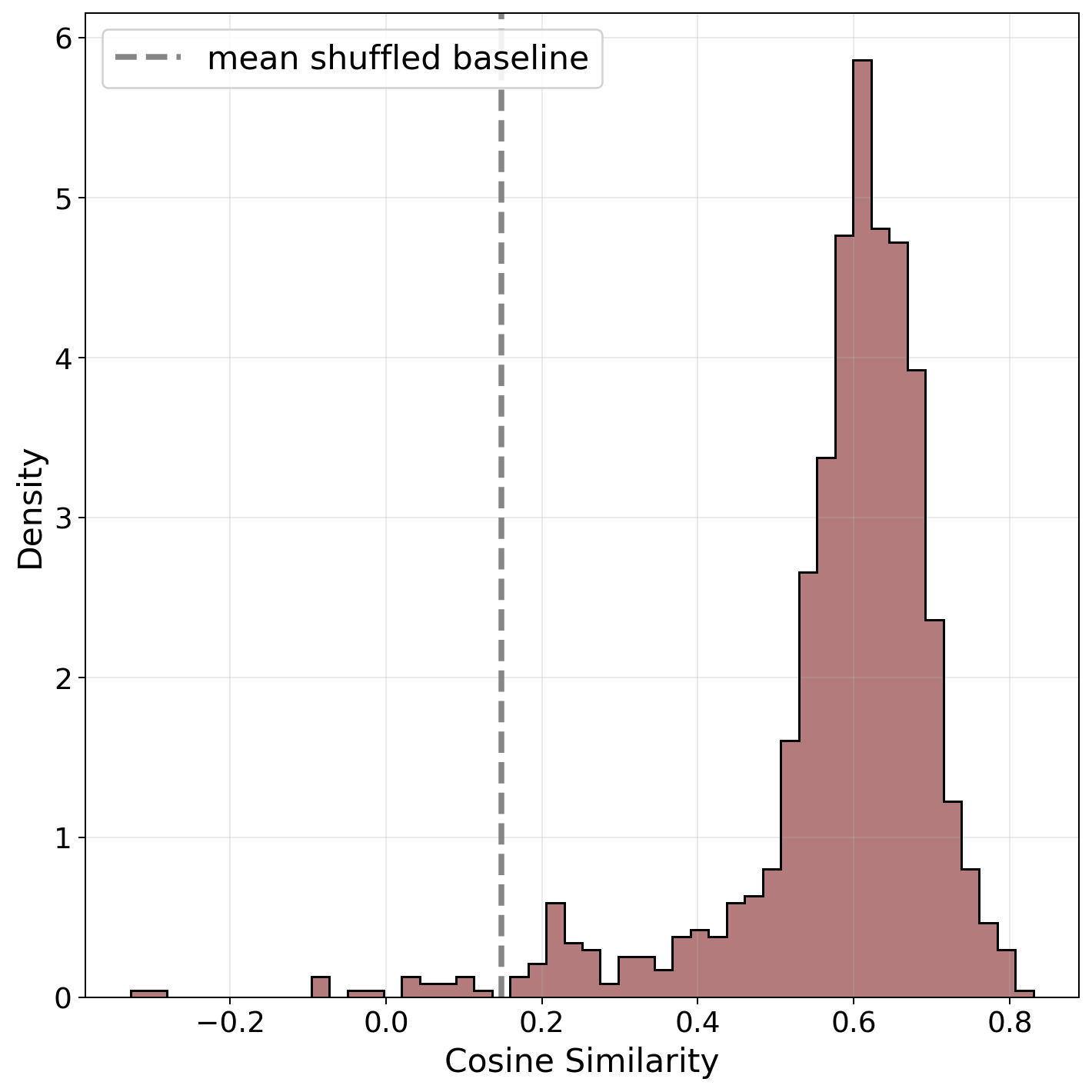}
    \label{fig:geometric_keys}
  \end{subfigure}
  \caption{\textbf{Feature composition and geometric alignment.} \textbf{(left)} TVD between student-space distributions of teacher neuron pairs, stratified by PMI. High-PMI pairs show lower TVD, indicating co-occurring features map to overlapping student regions. \textbf{(middle)} Cosine similarity between student value vectors and the teacher-distribution-weighted average of teacher value vectors, far above the shuffled baseline (${\approx}0.15$). \textbf{(right)} Same analysis for key vectors, confirming geometric alignment holds for both input and output projections.}
  \label{fig:cs2_composition}
  \vspace{-1em}
\end{figure*}

\subsection{Teacher--Student Neuron Mapping}
\label{sec:mapping}

To characterise how teacher neurons are encoded in the student, we associate each
teacher neuron $t_i$ with a distribution over student neurons via the conditional
expectation
\[
  a(s_j \mid t_i) \propto \mathbb{E}[s_j \mid t_i \text{ is  active} ],
\]
normalised to sum to one over all $N_S$ student neurons. The reverse mapping
$b(t_i \mid s_j)$ is defined analogously. The key question is whether these distributions are concentrated on a small number of neurons or spread uniformly. Concentration would imply structured reuse, while uniform spread would suggest no systematic alignment.
 
\subsection{Results}

\textbf{Quantifying polysemanticity.}
Figure~\ref{fig:cs2_mapping} (left) shows the cumulative mass functions for both mapping directions, with a random baseline confirming that the observed structure is specific to the trained student. The two directions are sharply asymmetric. In the forward direction $a(s_j \mid t_i)$, 80\% of a teacher neuron's mass concentrates on roughly 20 student neurons: each sparse feature maps to a localised region of the dense model rather than spreading uniformly. In the reverse direction $b(t_i \mid s_j)$, each student neuron draws 80\% of its mass from roughly 150 teacher neurons, consistent with the $8\times$ compression ratio ($20 \times 8 \approx 160$). This asymmetry is the expected signature of superposition: compression forces each dense neuron to absorb many sparse features, while each sparse feature retains a concentrated footprint in the dense representation.

\paragraph{Qualitative example.} Figure~\ref{fig:cs2_mapping} (right) shows student neuron $s_{310}$. Its four most influential teacher neurons encode entirely unrelated concepts: activating on \texttt{[own]} and propagating body-part tokens; activating on \texttt{[they]} and propagating verb tokens; activating on \texttt{[together]} and propagating \texttt{[again]}/\texttt{[in]}; and activating on \texttt{[upon]} and propagating time expressions. This is a direct, visually interpretable instance of polysemanticity.
 
\paragraph{Feature composition and absorption.} We ask whether teacher neurons that frequently co-occur (pointwise mutual information $\text{PMI} > 5$) map to similar regions of student space. Figure~\ref{fig:cs2_composition} (left) shows the distribution of Total Variation Distance (TVD) between the student-space distributions of teacher neuron pairs, stratified by PMI. High-PMI pairs show substantially lower TVD (mean ${\approx}0.29$) than anti-correlated pairs (mean ${\approx}0.44$). Co-occurring features are absorbed into the same dense neurons, mirroring the feature absorption and composition phenomena reported for SAEs \citep{chanin2026absorption, korznikov2025ortsae}.
 
\paragraph{Geometric alignment.}
We test whether a student neuron's weight vectors are a weighted average of its constituent teacher neurons' weight vectors. For each student neuron $s_j$, we calculate
\[
  \hat{\mathbf{w}}_j = \sum_{i=1}^{N_T} b(t_i \mid s_j)\,\mathbf{w}^{(t)}_i
  \label{eq:geometric}
\]
where $\mathbf{w}^{(t)}_i$ is either the key $\mathbf{k}^{(t)}_i$ or value $\mathbf{v}^{(t)}_i$ of teacher neuron $t_i$. The cosine similarity between $\hat{\mathbf{w}}_j$ and the true student weight vectors concentrates at 0.60, well above the shuffled baseline of $0.15$ (Figure~\ref{fig:cs2_composition}, middle and right). Thus, the weight structure of a polysemantic neuron is not arbitrary: it is a linear combination of the sparse features it encodes, providing a concrete mechanistic account of how compression shapes internal representations.
 
\section{Conclusion}
\label{sec:conclusion}

We introduced Sparse Transformer (ST), a language model with TopK-sparse MLPs, and found that this sparsity yields highly monosemantic neurons. Our primary model ($M{=}32$, $K{=}\text{8}$) reaches an AutoInterp score of 0.866 and supports compact, human-readable circuits. The two case studies exploit the known neuron basis of the ST to compare feature-learning methods and to trace how compression produces polysemanticity. An empirical power‑law relationship between width, sparsity, and loss further indicates that extreme sparsity can be offset by moderate width increases, suggesting a route toward scalable interpretable architectures.

Looking ahead, the ST testbed provides a controlled setting for evaluating interpretability methods against a known reference basis in natural-language models. We hope this facilitates systematic comparison of feature-learning techniques and empirical testing of theoretical claims regarding polysemanticity and superposition. We release all models and checkpoints to support such work.

\section*{Limitations}

Our models are trained on TinyStories and contain only 5 layers with a hidden
dimension of 256. This setting is sufficient to establish the ST as a
ground-truth testbed and to support interpretable circuits over concrete,
narrative-level concepts, but the model's limited capacity constrains the range
and abstraction of recoverable features. Neurons specialise in surface-level
syntactic patterns, character states, and simple narrative conventions; features
encoding abstract reasoning, world knowledge, or long-range dependencies lie
outside what this scale can represent. Our scaling results suggest a path to
addressing this directly, and training a larger ST along the identified
width--sparsity frontier is a natural direction for future work.

\bibliography{custom}

\appendix

\clearpage

\section{Robustness of Case Study \#1}
\label{app:cs1}

To test whether the SAE--Transcoder ranking depends on the sparsity mechanism or sparsity level, we additionally train both methods with $K{\in}\{4,16\}$ and with BatchTopK \citep{bussmann2024batchtopk} and Matryoshka \citep{bussmann2025learning} sparsity, each over 3 seeds (Table~\ref{tab:cs1robust}). We fix $D{=}N{=}8{,}192$ so that the dictionary can represent the true basis exactly and MMCS is comparable across methods. The Transcoder is better on every metric in every configuration.

\paragraph{Interpreting the gap.}
Let the ST MLP compute $\mathbf{y} = f(\mathbf{x}) = V_{\text{out}}\,\text{TopK}\bigl(\text{ReLU}(V_{\text{in}}\mathbf{x})\bigr)$. The SAE and the Transcoder minimise, respectively,
\begin{align}
\mathcal{L}_{\text{SAE}} &= \mathbb{E}\bigl\|\mathbf{y} - \mathbf{W}^{\text{dec}}\,\text{TopK}(\mathbf{W}^{\text{enc}}\mathbf{y})\bigr\|_2^2, \\
\mathcal{L}_{\text{TC}} &= \mathbb{E}\bigl\|\mathbf{y} - \mathbf{W}_{\text{out}}\,\text{TopK}(\mathbf{W}_{\text{in}}\mathbf{x})\bigr\|_2^2 .
\end{align}
From an optimisation standpoint, the Transcoder faces the harder problem. The SAE observes its target $\mathbf{y}$ directly and only needs to find a sparse code for it, whereas the Transcoder never observes $\mathbf{y}$ at its input: it must approximate the nonlinear map $f$ between two distinct spaces, which requires jointly identifying the keys $\mathbf{k}_i$ that determine the active set from $\mathbf{x}$ and the values $\mathbf{v}_i$ that produce $\mathbf{y}$. At the same time, the testbed realises the SAE's generative assumption exactly. By construction,
\[
\mathbf{y} = \sum_{i \in S(\mathbf{x})} a_i(\mathbf{x})\,\mathbf{v}_i, \qquad |S(\mathbf{x})| = K,\;\; a_i(\mathbf{x}) \geq 0,
\]
so every MLP output is a $K$-sparse non-negative combination of a fixed dictionary of $N$ vectors, and with $D{=}N$ the choice $\mathbf{W}^{\text{dec}}{=}V_{\text{out}}$ lies within the SAE's hypothesis class. This is the regime in which output reconstruction is best positioned to identify the true features, yet the SAE's value MMCS remains between $0.365$ and $0.469$ across all configurations. Finally, the Transcoder's advantage does not rest on matching the ST MLP architecture. With $K{\in}\{4,16\}$, BatchTopK, or Matryoshka sparsity, its gating differs from the ST's per-token TopK with $K{=}8$, so it no longer shares the functional form of $f$. Nevertheless, it outperforms the SAE on every metric in each of these configurations, with value MMCS at least $1.5\times$ higher in all of them (e.g., $0.661$ vs.\ $0.378$ at $K{=}4$; $0.701$ vs.\ $0.411$ with BatchTopK). The gap therefore persists whether or not the Transcoder can represent $f$ exactly, which points to a genuine difference between reconstructing a representation and approximating a computation.

\section{Circuit Metrics and Additional Circuits}
\label{app:circuits}

To test whether the circuit in Section~\ref{sec:circuit} is representative, we extract circuits for 15 additional prompts in the style of TinyStories, all with target token probability $P_0 \geq 0.80$. For each prompt, we apply the same beam-search ablation protocol with a budget of 50 neurons and record $F$ and $C$ at every circuit size $n \in [1, 50]$. Table~\ref{tab:circuit-per-prompt} reports per-prompt metrics, and Table~\ref{tab:circuit-grid} aggregates them by circuit size.

\paragraph{Reading the metrics.} $C$ is defined such that lower is better: $C{=}0.2$ means that ablating the circuit reduces the target probability fivefold. As a reference, ablating 11 random neurons leaves $P(\texttt{[eat]})$ virtually unchanged ($C{\approx}0.998$). $F$ can exceed $1$ when the neurons outside the circuit are net-suppressive; the circuit then reproduces the net effect of the full model rather than its exact probability.

\paragraph{Results.} A median of 12 neurons suffices to reach $F \geq 0.95$, achieved by 15 of 16 prompts within the budget. At a fixed budget of 20 neurons, $F$ averages $1.01$ and $C$ has a median of $0.12$; 14 of 16 circuits reach $F \geq 0.95$ and 11 of 16 reach $C \leq 0.2$ at this size. Faithfulness saturates faster than completeness: for some prompts (e.g., \texttt{[mirror]}, \texttt{[read]}), a small circuit reproduces the prediction while its ablation leaves much of the target probability intact, indicating redundant pathways that support the same token. Overall, compact circuits are a stable property of the model across prediction probabilities ($81.5\%$--$99.9\%$) and semantic contexts. Figures 6--20 show the $F$ and $C$ curves for each prompt..


\FloatBarrier




\section{Neuron Dashboards}
\label{app:dashboards}

This appendix presents dashboards for 20 randomly sampled neurons (4 per layer). For each neuron, we display its top activating contexts at multiple intensity tiers. To aid interpretation, we also project the input weight vector $W_{\text{in}}$ onto the embedding space and the output weight vector $W_{\text{out}}$ onto the unembedding space-a Logit Lens \citep{nostalgebraist2020logit} applied to individual MLP neurons. The resulting top positive and negative logits reveal what patterns the neuron detects and which tokens it promotes or suppresses. Together, these examples illustrate that ST neurons encode clear, specific concepts.

\clearpage

\begin{table*}[h!]
  \centering
  \small
  \begin{tabular}{llccc}
    \toprule
    \textbf{Setting} & \textbf{Method} & \textbf{NMSE} $\downarrow$ & \textbf{MMCS} enc./key $\uparrow$ & \textbf{MMCS} dec./value $\uparrow$ \\
    \midrule
    TopK ($K{=}8$)       & Transcoder & $0.0094 \pm 0.0005$ & $0.564 \pm 0.0002$ & $0.709 \pm 0.0001$ \\
                         & SAE        & $0.0358 \pm 0.0020$ & $0.410 \pm 0.0001$ & $0.464 \pm 0.0001$ \\
    \midrule
    TopK ($K{=}4$)       & Transcoder & $0.0237 \pm 0.0012$ & $0.493 \pm 0.0002$ & $0.661 \pm 0.0001$ \\
                         & SAE        & $0.0681 \pm 0.0037$ & $0.340 \pm 0.0000$ & $0.378 \pm 0.0000$ \\
    \midrule
    TopK ($K{=}16$)      & Transcoder & $0.0068 \pm 0.0004$ & $0.593 \pm 0.0001$ & $0.720 \pm 0.0001$ \\
                         & SAE        & $0.0247 \pm 0.0013$ & $0.415 \pm 0.0001$ & $0.469 \pm 0.0002$ \\
    \midrule
    BatchTopK ($K{=}8$)  & Transcoder & $0.0099 \pm 0.0005$ & $0.564 \pm 0.0002$ & $0.701 \pm 0.0001$ \\
                         & SAE        & $0.0451 \pm 0.0025$ & $0.362 \pm 0.0001$ & $0.411 \pm 0.0001$ \\
    \midrule
    Matryoshka ($K{=}8$) & Transcoder & $0.0361 \pm 0.0017$ & $0.422 \pm 0.0001$ & $0.559 \pm 0.0001$ \\
                         & SAE        & $0.0528 \pm 0.0029$ & $0.314 \pm 0.0001$ & $0.365 \pm 0.0001$ \\
    \bottomrule
  \end{tabular}
  \caption{\textbf{SAE vs.\ Transcoder under alternative sparsity settings} (mean $\pm$ std over 3 seeds). As in Section~\ref{sec:cs1}, SAE encoder and decoder vectors (``enc.'', ``dec.'') are matched against the true value vectors, whereas Transcoder input and output vectors (``key'', ``value'') are matched against the true keys and values, respectively.}
  \label{tab:cs1robust}
\end{table*}

\begin{table*}[t!]
\centering
\small
\begin{tabular}{rlccccccc}
\toprule
\# & Target & $P_0$ (\%) & $n_F$ & $F(n_F)$ & $C(n_F)$ & $n_C$ & $F(20)$ & $C(20)$ \\
\midrule
1  & \texttt{[eat]}      & 98.3 & 11 & 0.956 & 0.198 & 11 & 1.010 & 0.142 \\
2  & \texttt{[hug]}      & 89.1 & 12 & 0.961 & 0.114 & 9  & 1.082 & 0.063 \\
3  & \texttt{[chase]}    & 81.5 & 13 & 0.968 & 0.467 & 39 & 1.112 & 0.338 \\
4  & \texttt{[prince]}   & 96.9 & 14 & 0.951 & 0.139 & 11 & 0.982 & 0.085 \\
5  & \texttt{[voice]}    & 94.8 & 5  & 0.986 & 0.331 & 6  & 1.054 & 0.057 \\
6  & \texttt{[day]}      & 99.1 & 12 & 0.954 & 0.149 & 10 & 0.996 & 0.089 \\
7  & \texttt{[treasure]} & 91.7 & 12 & 0.954 & 0.606 & 37 & 1.027 & 0.379 \\
8  & \texttt{[table]}    & 98.8 & 19 & 0.961 & 0.008 & 4  & 0.968 & 0.008 \\
9  & \texttt{[door]}     & 99.9 & 49 & 0.952 & 0.016 & 27 & 0.906 & 0.711 \\
10 & \texttt{[read]}     & 88.8 & 10 & 0.961 & 0.596 & -- & 0.976 & 0.378 \\
11 & \texttt{[paper]}    & 97.7 & -- & --    & --    & 18 & 0.939 & 0.166 \\
12 & \texttt{[better]}   & 99.7 & 10 & 0.973 & 0.044 & 9  & 1.000 & 0.003 \\
13 & \texttt{[mirror]}   & 99.9 & 3  & 0.961 & 0.905 & -- & 1.000 & 0.625 \\
14 & \texttt{[oven]}     & 99.6 & 14 & 0.961 & 0.055 & 10 & 1.002 & 0.018 \\
15 & \texttt{[cry]}      & 83.8 & 15 & 0.989 & 0.009 & 3  & 1.041 & 0.006 \\
16 & \texttt{[tall]}     & 87.2 & 8  & 0.953 & 0.534 & 20 & 1.075 & 0.150 \\
\midrule
\multicolumn{2}{l}{Median [min, max]} & 97.3 & 12 [3, 49] & -- & -- & 10.5 [3, 39] & 1.00 & 0.12 \\
\bottomrule
\end{tabular}
\caption{\textbf{Per-prompt circuit metrics.} $P_0$ is the full-model probability of the target token. $n_F$ is the smallest circuit size with $F \geq 0.95$ (reached by 15 of 16 prompts), and $F(n_F)$, $C(n_F)$ are the metrics at that size; $n_C$ is the smallest size with $C \leq 0.2$ (reached by 14 of 16 prompts); $F(20)$, $C(20)$ are the metrics at a fixed budget of 20 neurons. Entries marked -- are not reached within the 50-neuron budget; medians of $n_F$ and $n_C$ are over prompts that reach the threshold. Row 1 is the circuit in Figure~\ref{fig:circuit}.}
\label{tab:circuit-per-prompt}
\end{table*}

\begin{table*}[t!]
\centering
\small
\setlength{\tabcolsep}{4pt}
\begin{tabular}{lcccccccccccc}
\toprule
Neurons $n$ & 1 & 2 & 3 & 5 & 8 & 10 & 15 & 20 & 25 & 30 & 40 & 50 \\
\midrule
$F$ mean   & 0.11 & 0.21 & 0.30 & 0.49 & 0.74 & 0.85 & 0.98 & 1.01 & 1.02 & 1.03 & 1.03 & 1.03 \\
$F$ std    & 0.16 & 0.25 & 0.29 & 0.29 & 0.22 & 0.16 & 0.06 & 0.05 & 0.05 & 0.06 & 0.06 & 0.06 \\
$F$ median & 0.04 & 0.09 & 0.16 & 0.47 & 0.82 & 0.90 & 0.99 & 1.00 & 1.00 & 1.00 & 1.01 & 1.01 \\
\midrule
$C$ mean   & 0.92 & 0.84 & 0.76 & 0.62 & 0.45 & 0.35 & 0.26 & 0.20 & 0.17 & 0.12 & 0.09 & 0.07 \\
$C$ std    & 0.14 & 0.18 & 0.23 & 0.24 & 0.26 & 0.28 & 0.25 & 0.22 & 0.17 & 0.12 & 0.09 & 0.07 \\
$C$ median & 0.96 & 0.89 & 0.86 & 0.66 & 0.41 & 0.26 & 0.19 & 0.12 & 0.11 & 0.08 & 0.06 & 0.06 \\
\bottomrule
\end{tabular}
\caption{\textbf{Faithfulness ($F$, higher is better) and completeness ($C$, lower is better) as a function of circuit size $n$}, aggregated over the 16 prompts (population std). Per-prompt values for every $n \in [1, 50]$ are in the released repository.}
\label{tab:circuit-grid}
\end{table*}

\clearpage

\begin{figure}[h!]
\centering
\includegraphics[width=\columnwidth]{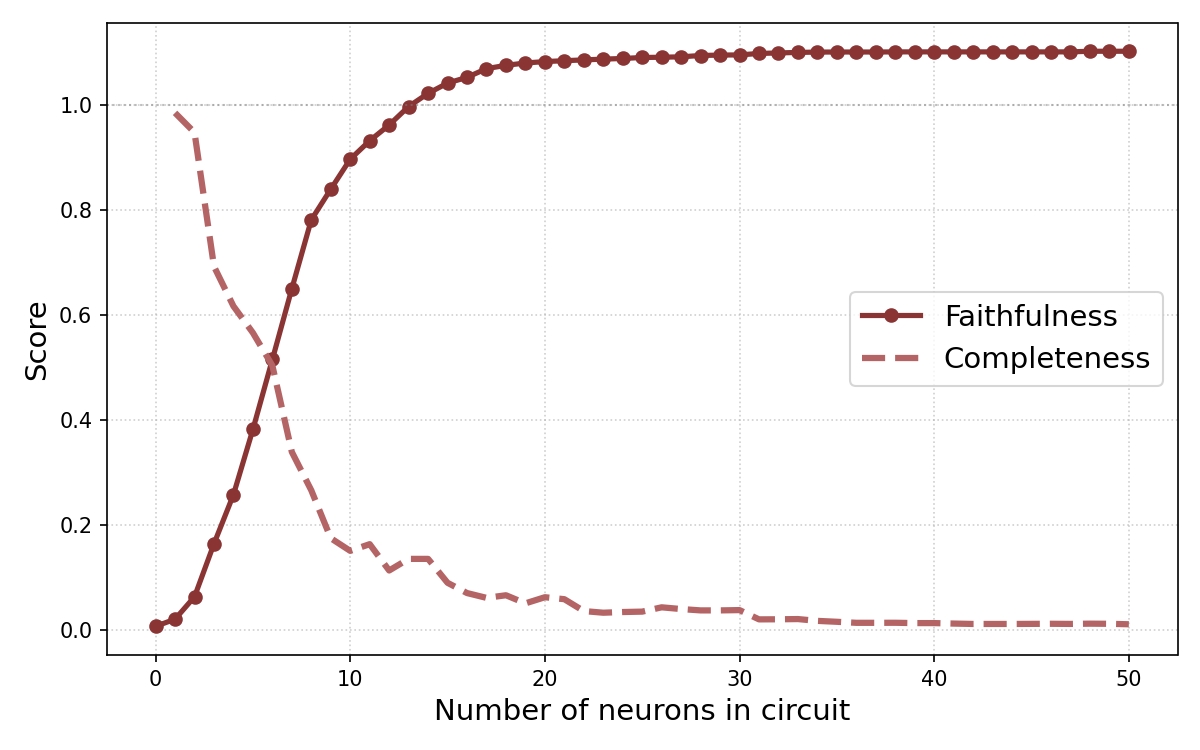}
\caption{\textbf{Circuit for \emph{``She felt very sad so her friend gave her a big warm''} $\to$ \texttt{[hug]} ($p{=}89.1\%$).}}
\end{figure}

\begin{figure}[h!]
\centering
\includegraphics[width=\columnwidth]{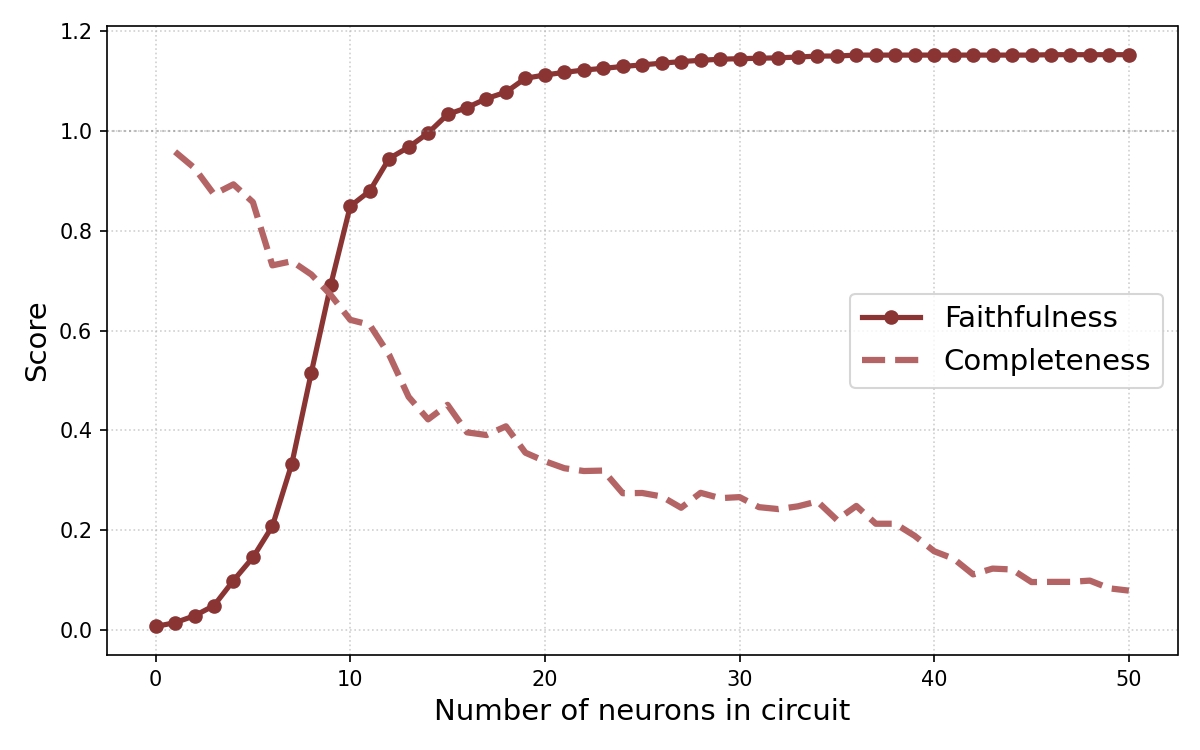}
\caption{\textbf{Circuit for \emph{``The cat saw a mouse and started to''} $\to$ \texttt{[chase]} ($p{=}81.5\%$).}}
\end{figure}

\begin{figure}[h!]
\centering
\includegraphics[width=\columnwidth]{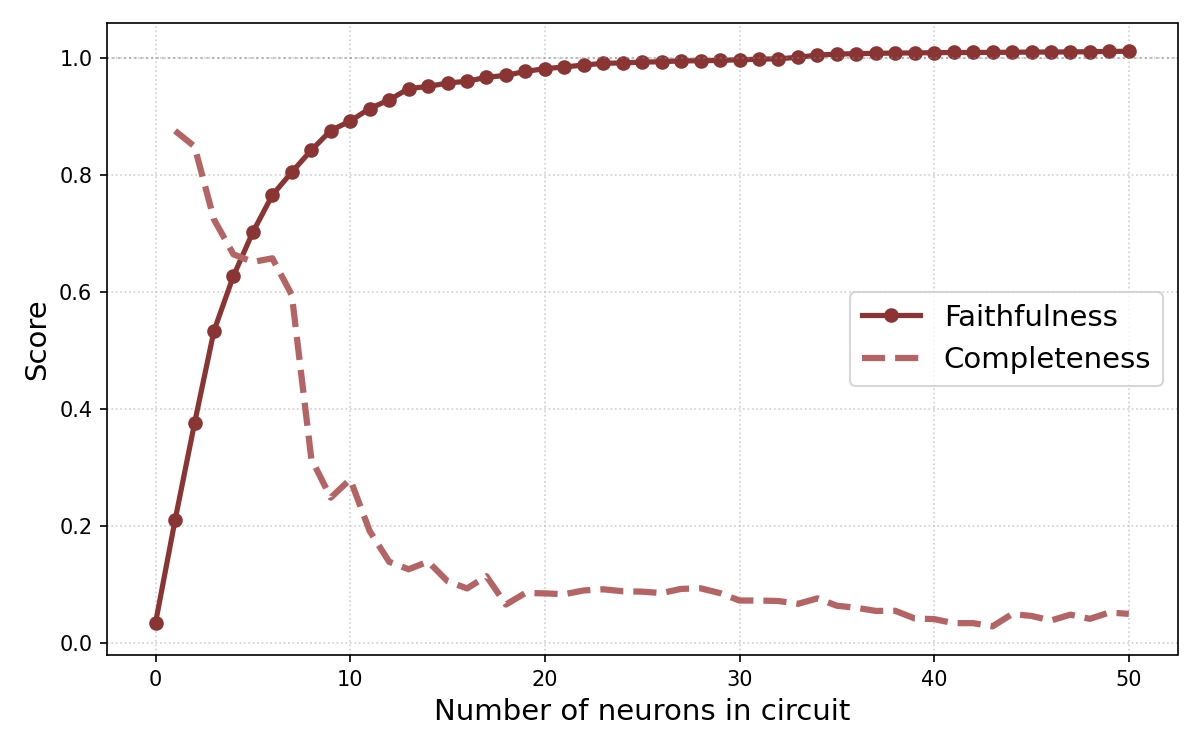}
\caption{\textbf{Circuit for \emph{``The princess kissed the frog and it turned into a handsome''} $\to$ \texttt{[prince]} ($p{=}96.9\%$).}}
\end{figure}

\begin{figure}[h!]
\centering
\includegraphics[width=\columnwidth]{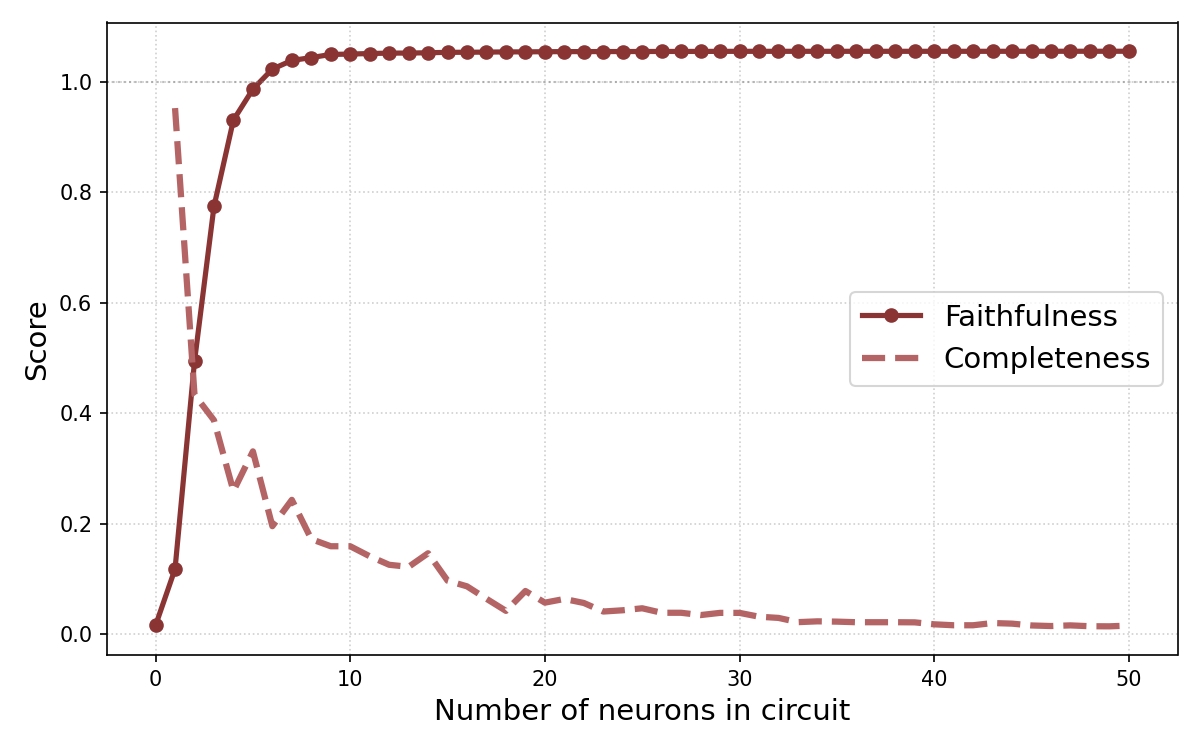}
\caption{\textbf{Circuit for \emph{``The robot beeped and said in a friendly''} $\to$ \texttt{[voice]} ($p{=}94.8\%$).}}
\end{figure}

\begin{figure}[h!]
\centering
\includegraphics[width=\columnwidth]{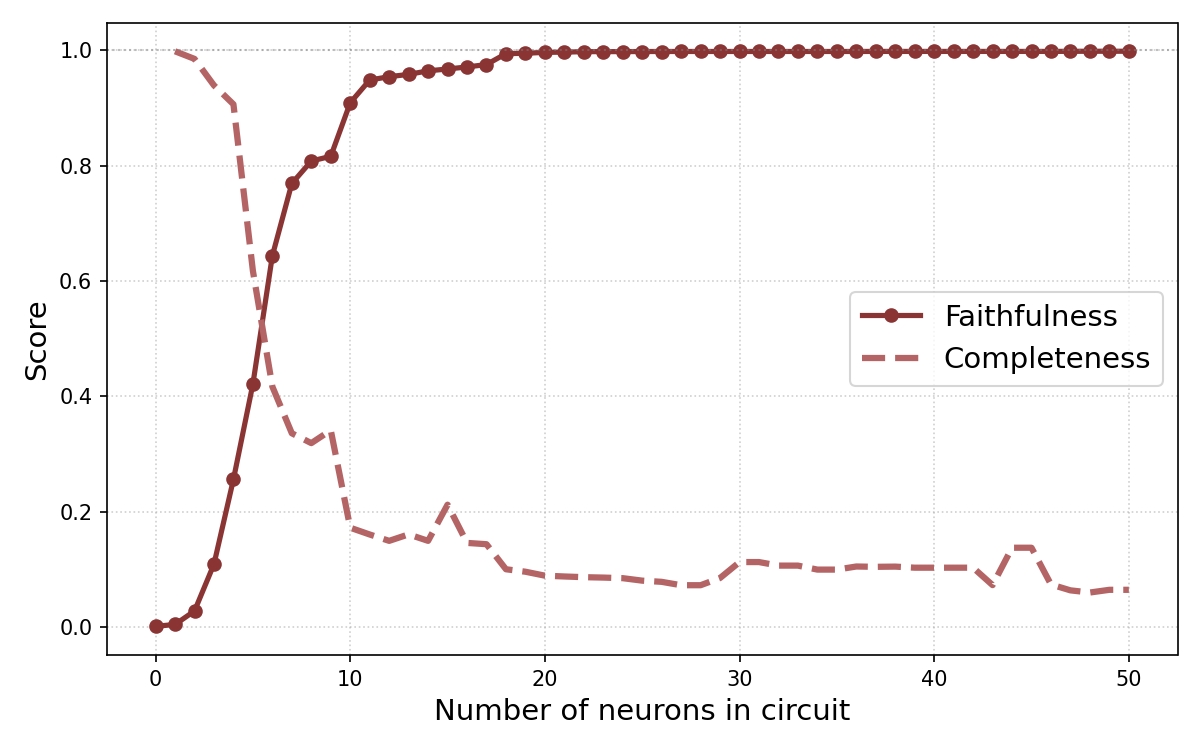}
\caption{\textbf{Circuit for \emph{``She planted a seed and watered it every''} $\to$ \texttt{[day]} ($p{=}99.1\%$).}}
\end{figure}

\begin{figure}[h!]
\centering
\includegraphics[width=\columnwidth]{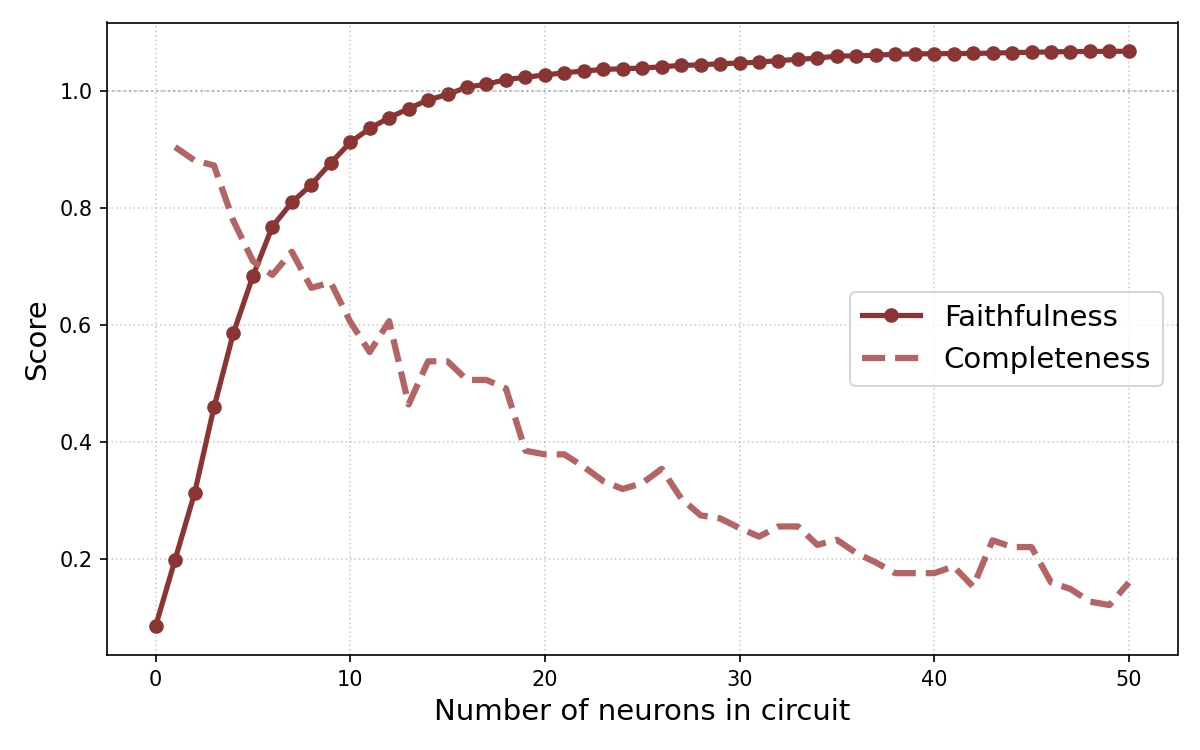}
\caption{\textbf{Circuit for \emph{``The pirate looked at the map to find the hidden''} $\to$ \texttt{[treasure]} ($p{=}91.7\%$).}}
\end{figure}

\begin{figure}[h!]
\centering
\includegraphics[width=\columnwidth]{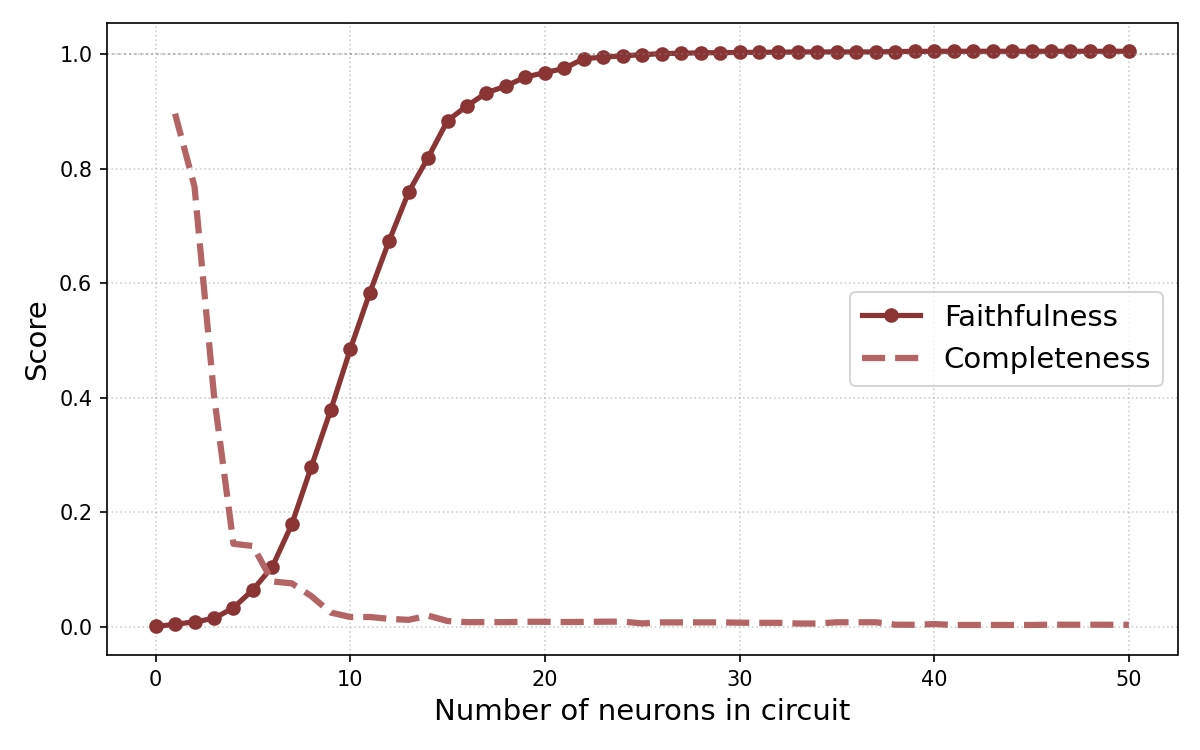}
\caption{\textbf{Circuit for \emph{``He washed his hands before sitting at the dinner''} $\to$ \texttt{[table]} ($p{=}98.8\%$).}}
\end{figure}

\begin{figure}[h!]
\centering
\includegraphics[width=\columnwidth]{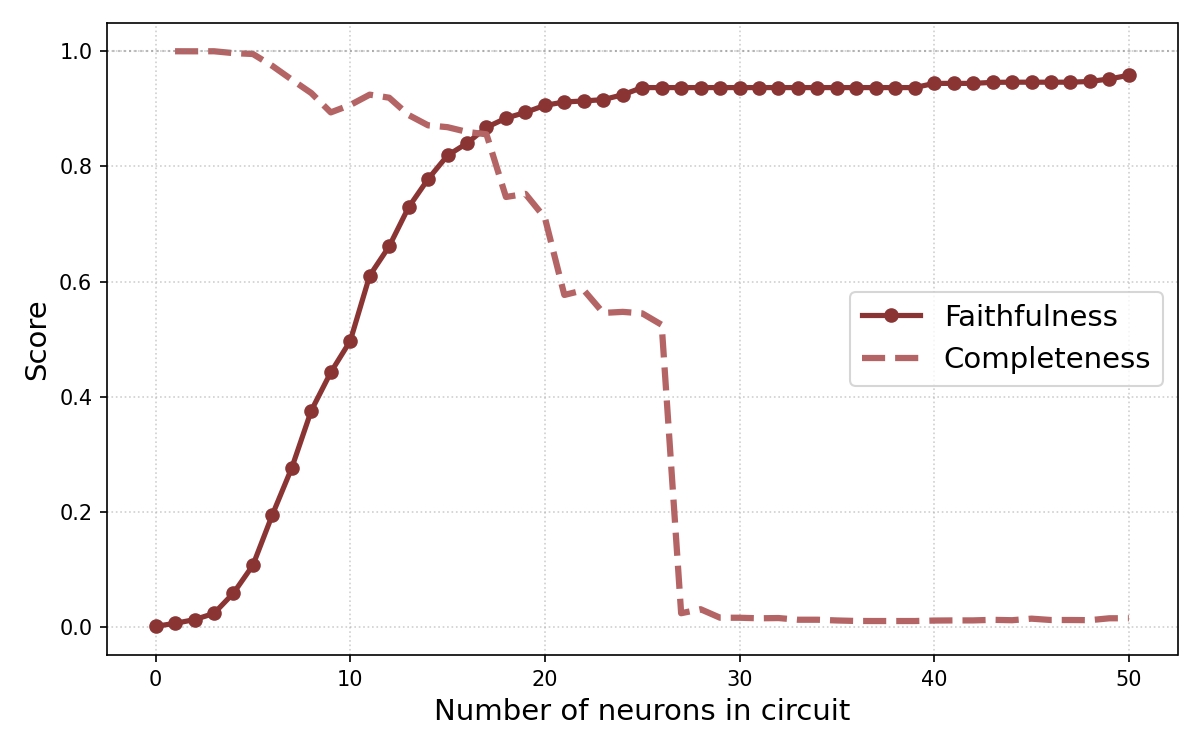}
\caption{\textbf{Circuit for \emph{``She put on her shoes and ran out the''} $\to$ \texttt{[door]} ($p{=}99.9\%$).}}
\end{figure}

\begin{figure}[h!]
\centering
\includegraphics[width=\columnwidth]{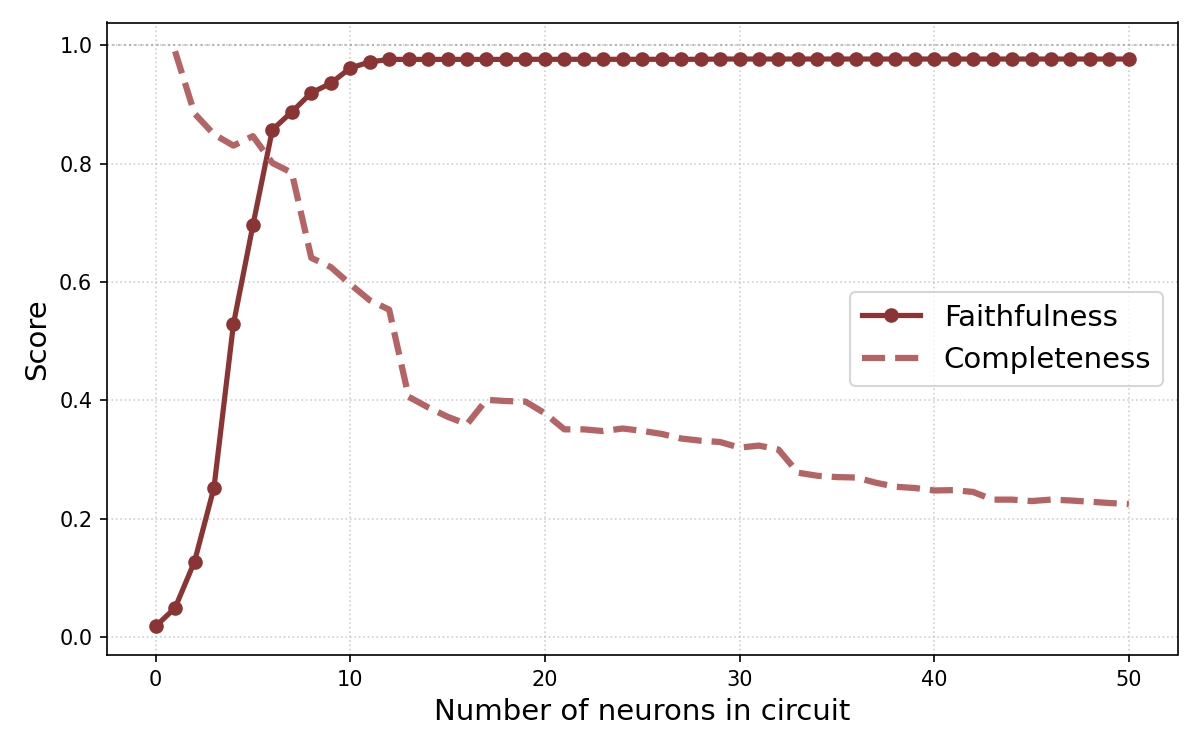}
\caption{\textbf{Circuit for \emph{``She opened the book and started to''} $\to$ \texttt{[read]} ($p{=}88.8\%$).}}
\end{figure}

\begin{figure}[h!]
\centering
\includegraphics[width=\columnwidth]{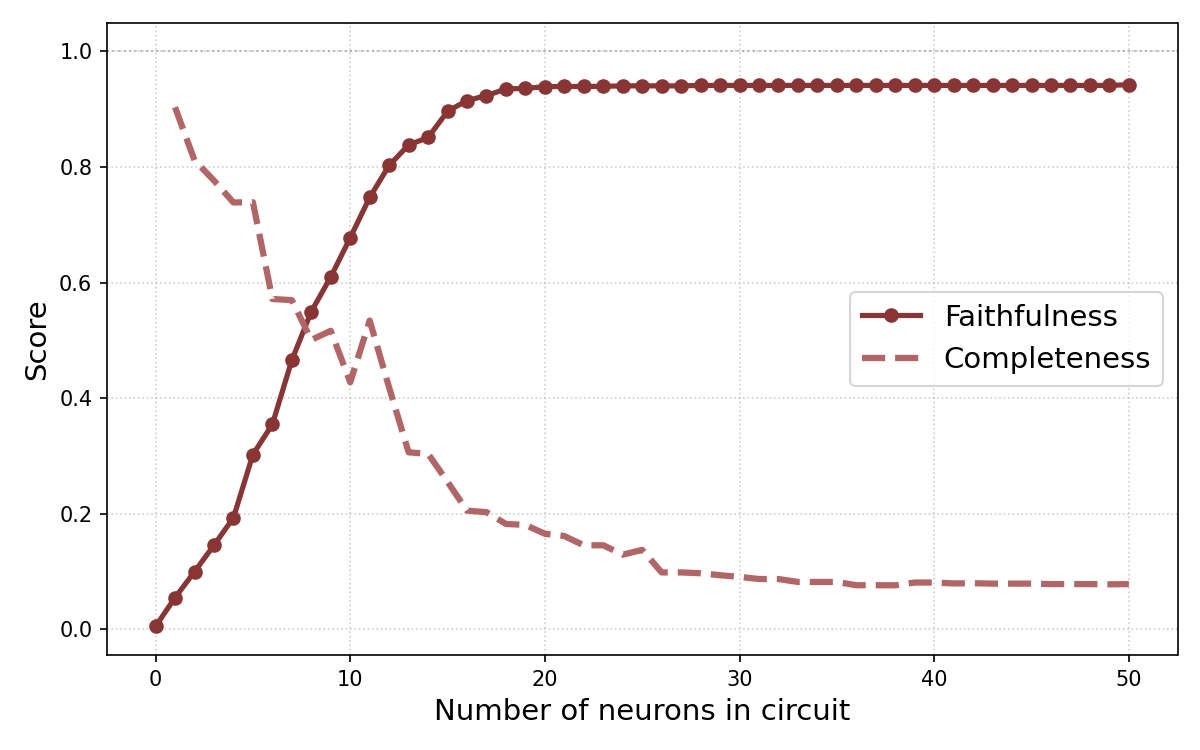}
\caption{\textbf{Circuit for \emph{``He wrapped the gift in colorful shiny''} $\to$ \texttt{[paper]} ($p{=}97.7\%$).}}
\end{figure}

\begin{figure}[h!]
\centering
\includegraphics[width=\columnwidth]{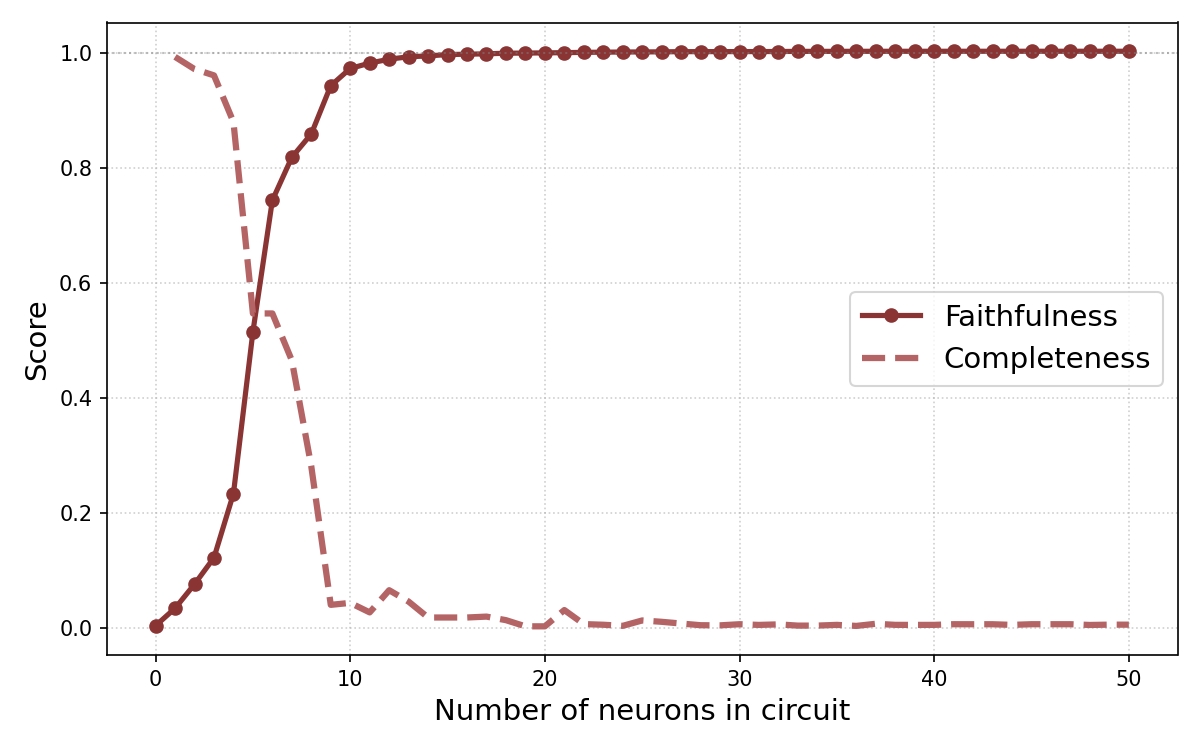}
\caption{\textbf{Circuit for \emph{``She drank the hot soup and felt much''} $\to$ \texttt{[better]} ($p{=}99.7\%$).}}
\end{figure}

\begin{figure}[h!]
\centering
\includegraphics[width=\columnwidth]{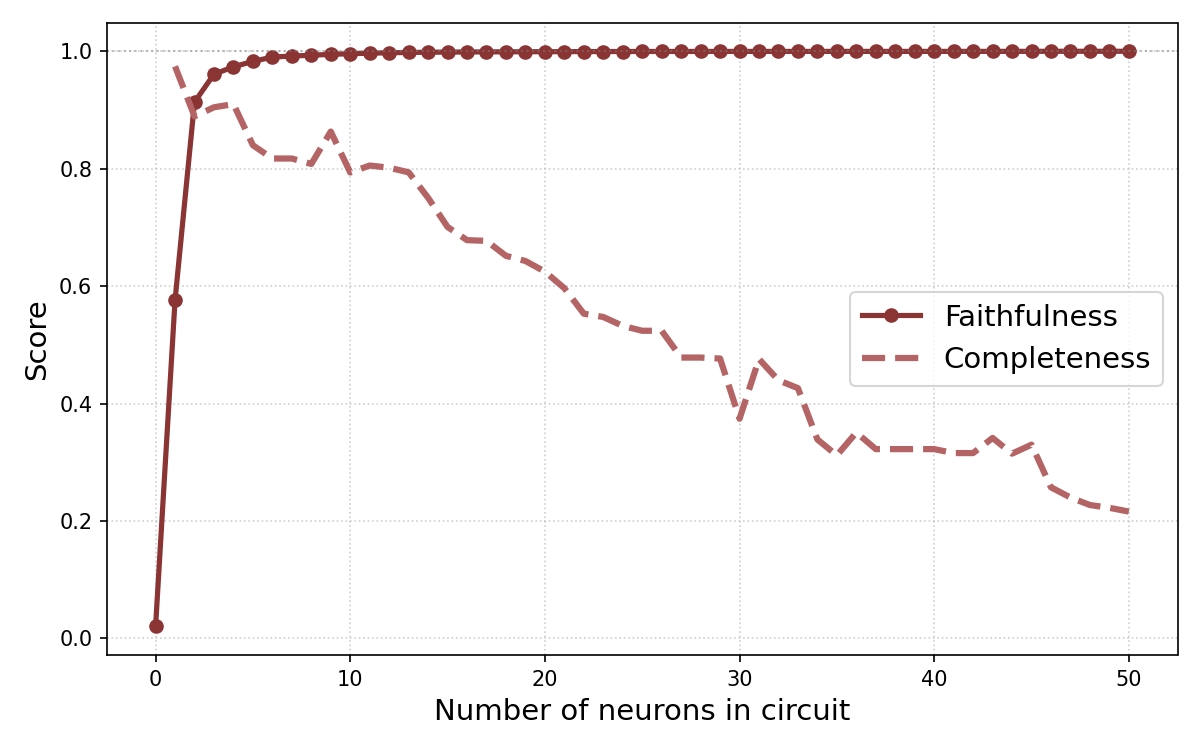}
\caption{\textbf{Circuit for \emph{``She brushed her hair and looked in the''} $\to$ \texttt{[mirror]} ($p{=}99.9\%$).}}
\end{figure}

\begin{figure}[h!]
\centering
\includegraphics[width=\columnwidth]{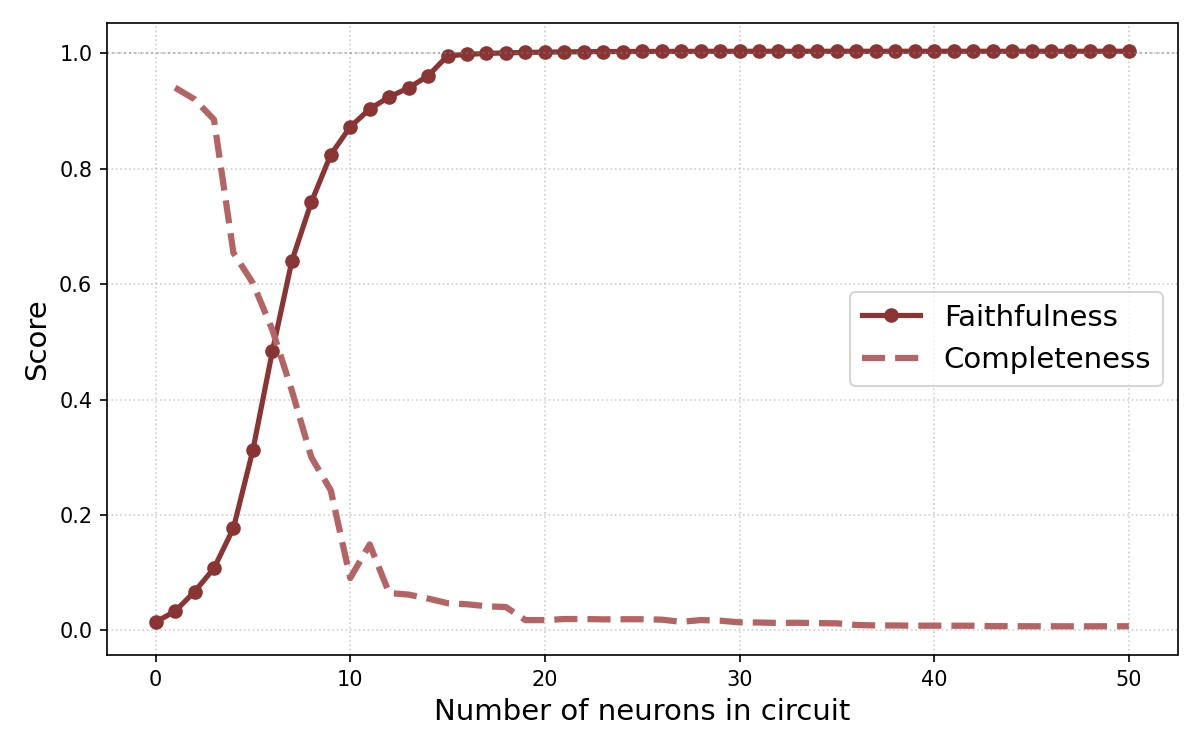}
\caption{\textbf{Circuit for \emph{``She baked a cake and put it in the''} $\to$ \texttt{[oven]} ($p{=}99.6\%$).}}
\end{figure}

\begin{figure}[h!]
\centering
\includegraphics[width=\columnwidth]{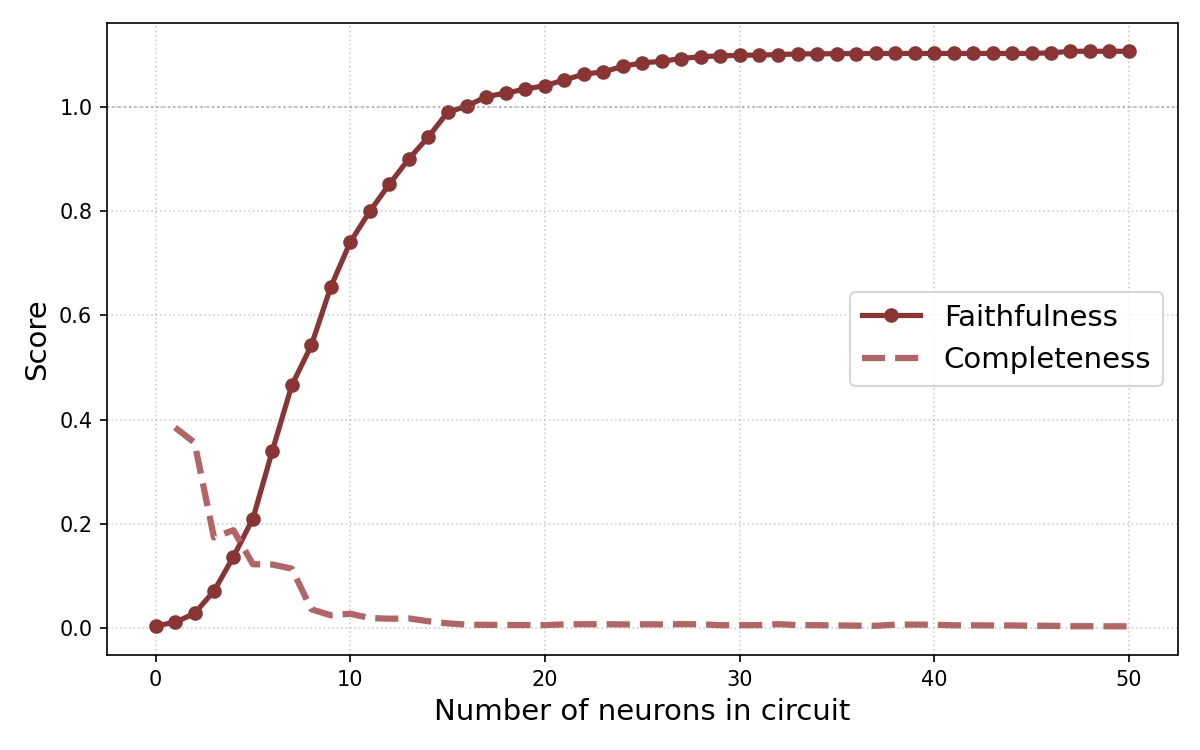}
\caption{\textbf{Circuit for \emph{``The baby dropped her toy and started to''} $\to$ \texttt{[cry]} ($p{=}83.8\%$).}}
\end{figure}

\begin{figure}[h!]
\centering
\includegraphics[width=\columnwidth]{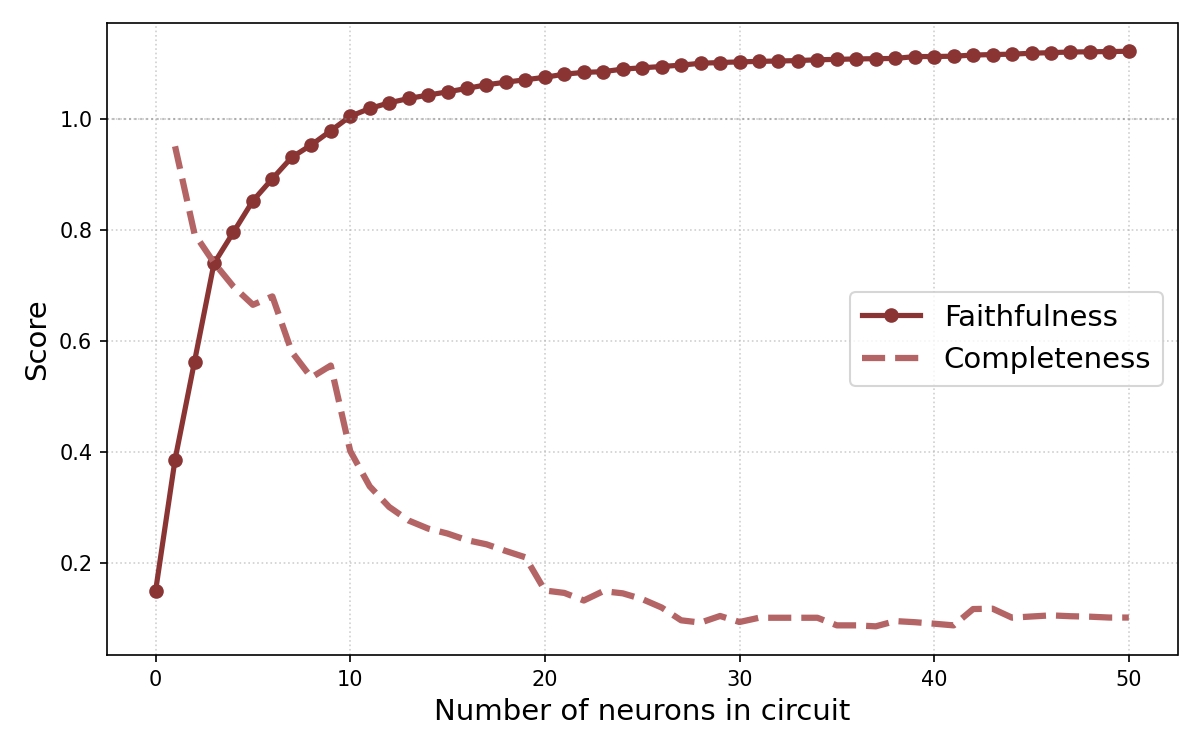}
\caption{\textbf{Circuit for \emph{``He stacked the blocks until the tower was very''} $\to$ \texttt{[tall]} ($p{=}87.2\%$).}}
\end{figure}

\clearpage
\begin{figure*}[h!]
\centering
\includegraphics[width=\textwidth]{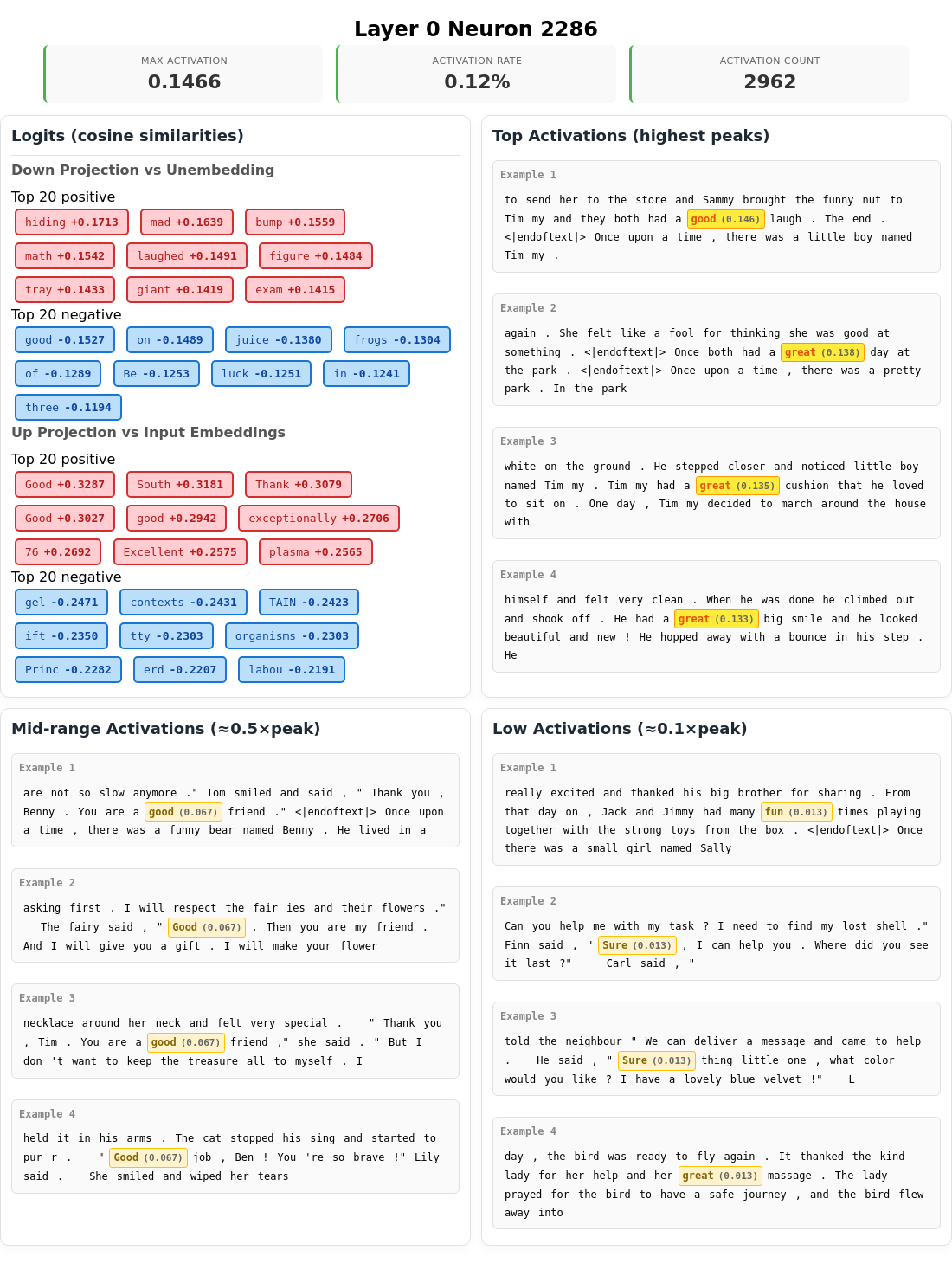}
\caption{\textbf{L0‑N2286} activates on positive adjectives (\emph{good, great}) and friendly interpersonal contexts.}
\end{figure*}

\clearpage
\begin{figure*}[h!]
\centering
\includegraphics[width=\textwidth]{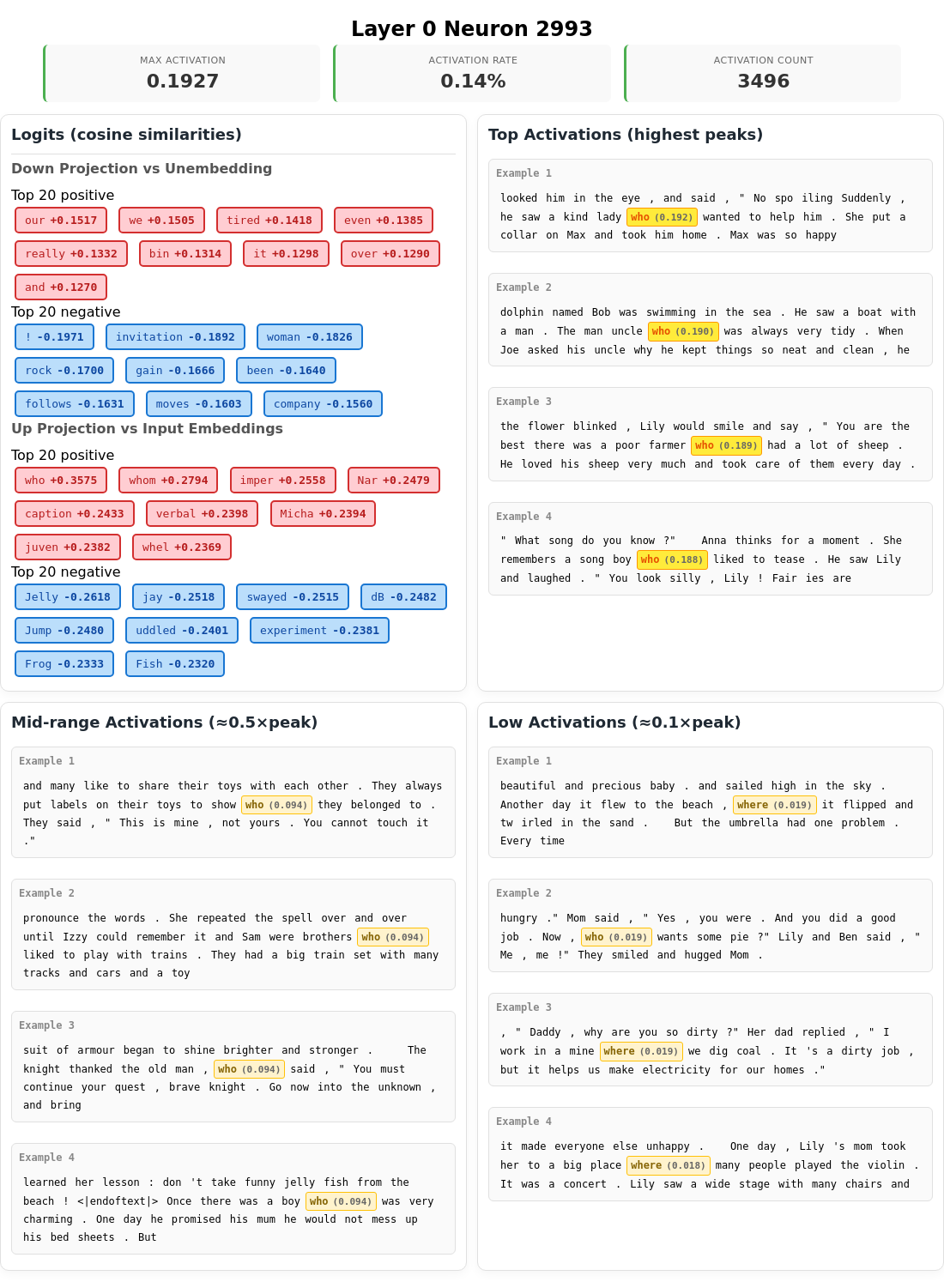}
\caption{\textbf{L0‑N2993} detects the relative pronoun \emph{who} (and its variants \emph{whom}) in narrative contexts, particularly after a character introduction.}
\end{figure*}

\clearpage
\begin{figure*}[h!]
\centering
\includegraphics[width=\textwidth]{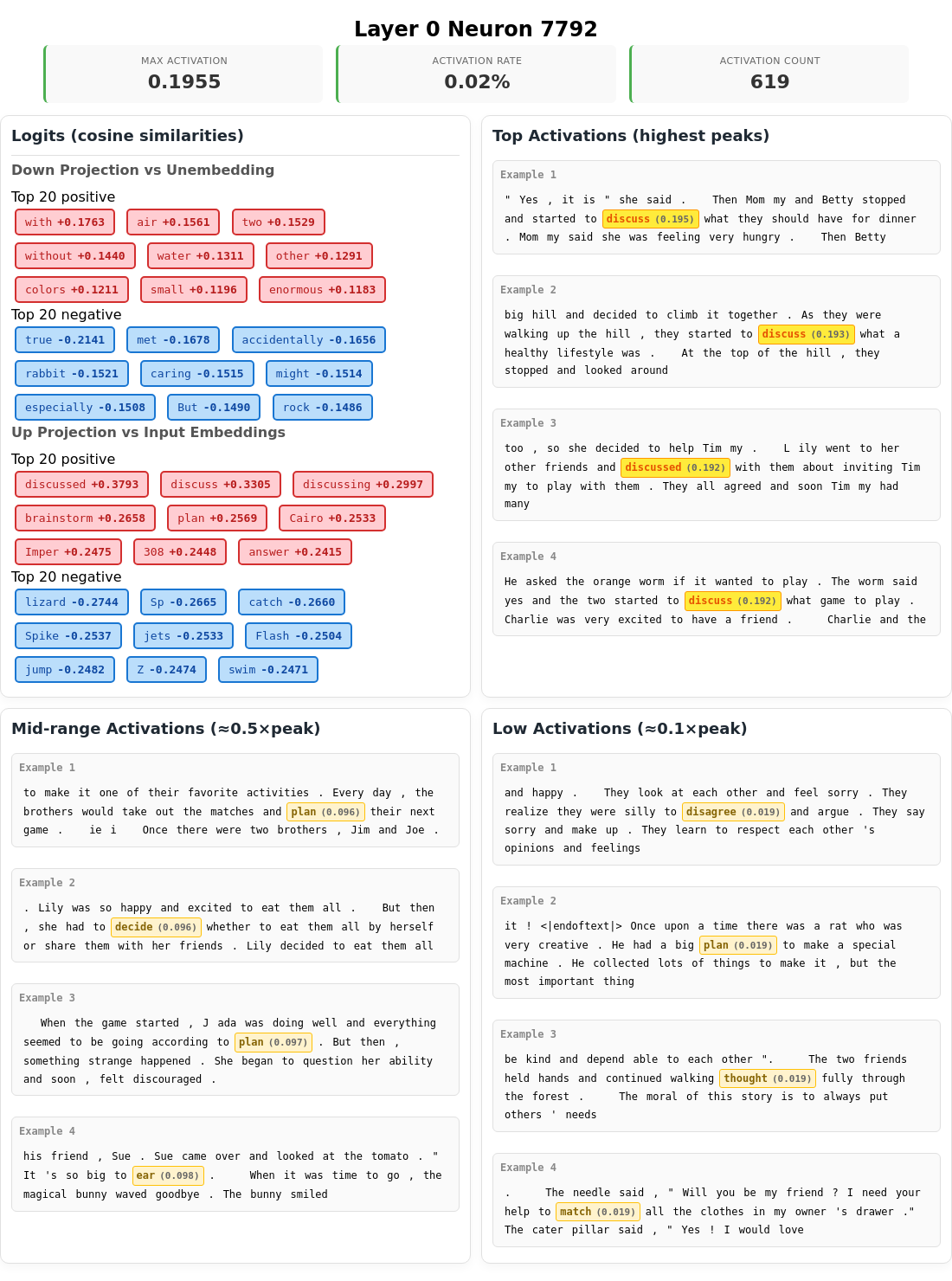}
\caption{\textbf{L0‑N7792} responds to planning and discussion verbs (\emph{discussed, brainstorm, plan}).}
\end{figure*}

\clearpage
\begin{figure*}[h!]
\centering
\includegraphics[width=\textwidth]{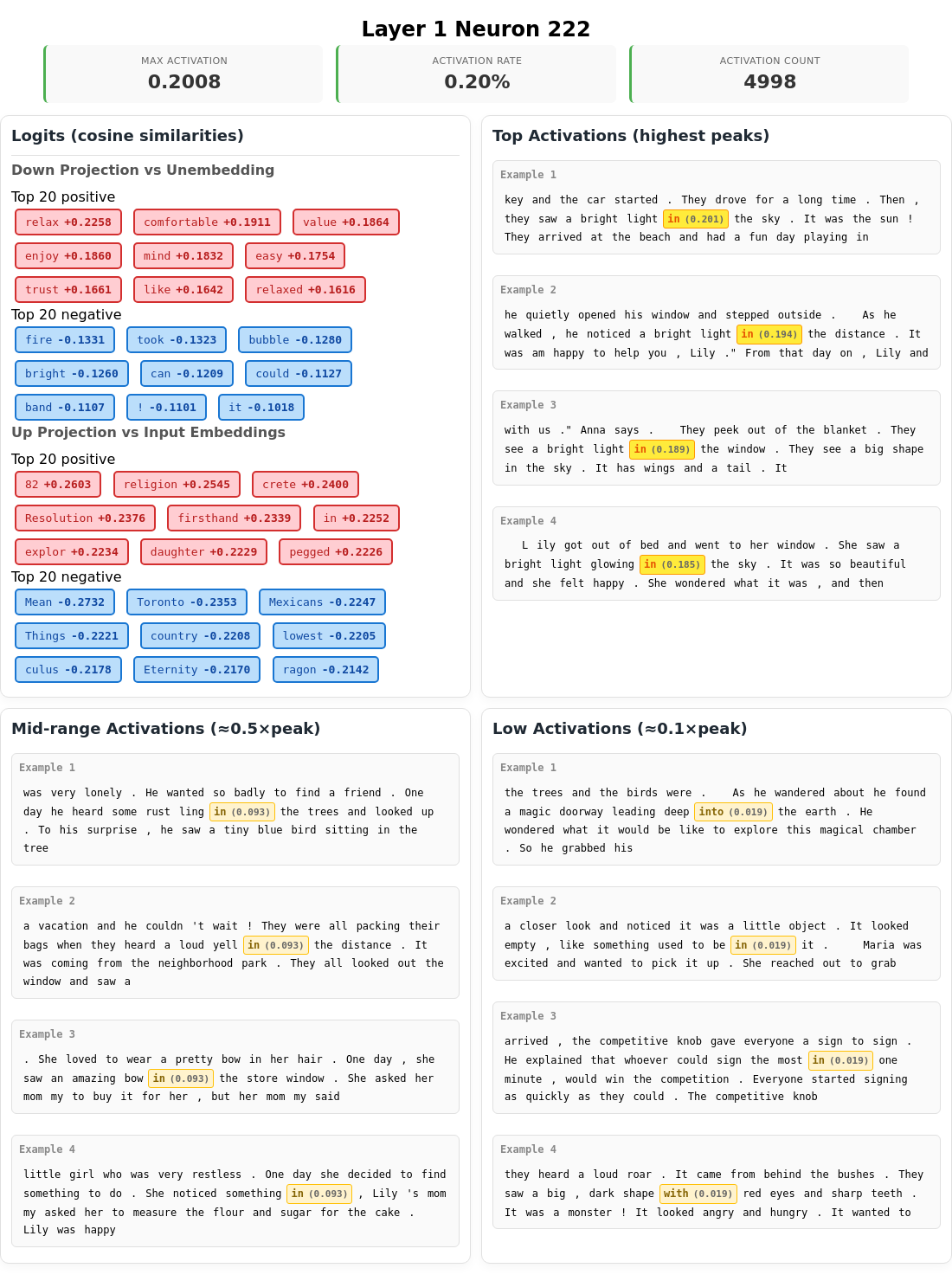}
\caption{\textbf{L1‑N222} activates on the preposition \emph{in} in contexts where a character notices or discovers something in their surroundings (e.g.\ \emph{``a bright light in the sky''}, \emph{``something in the distance''}). Its output weight vector promotes calm, positive continuations (\emph{relax, comfortable, enjoy, trust}).}
\end{figure*}

\clearpage
\begin{figure*}[h!]
\centering
\includegraphics[width=\textwidth]{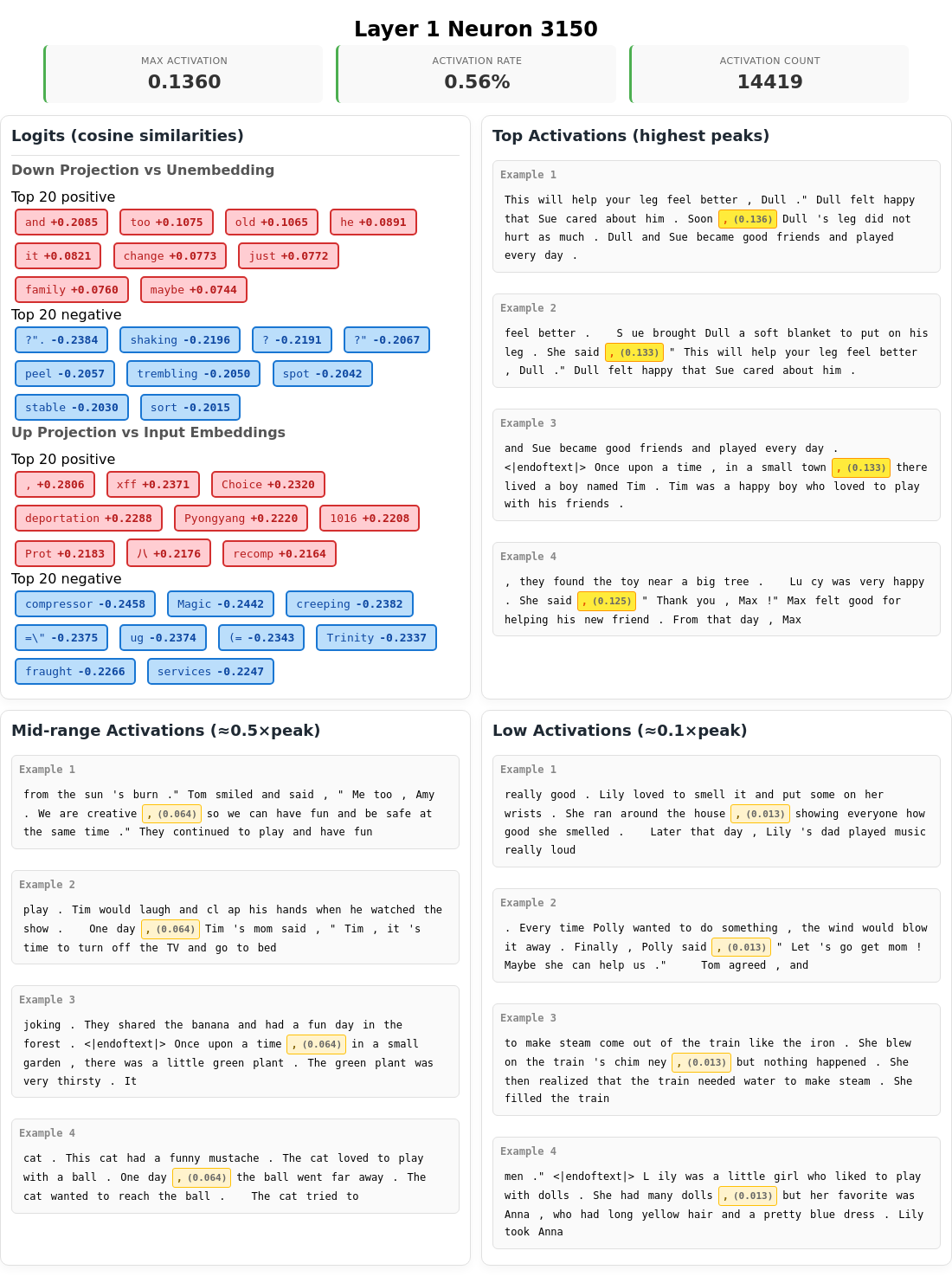}
\caption{\textbf{L1‑N3150} activates on the comma following a character's speech or completed action, marking a narrative continuation point. Its output weight vector promotes continuation tokens (\emph{and, he, it, change}), suggesting it signals that another clause or reaction is about to follow.}
\end{figure*}

\clearpage
\begin{figure*}[h!]
\centering
\includegraphics[width=\textwidth]{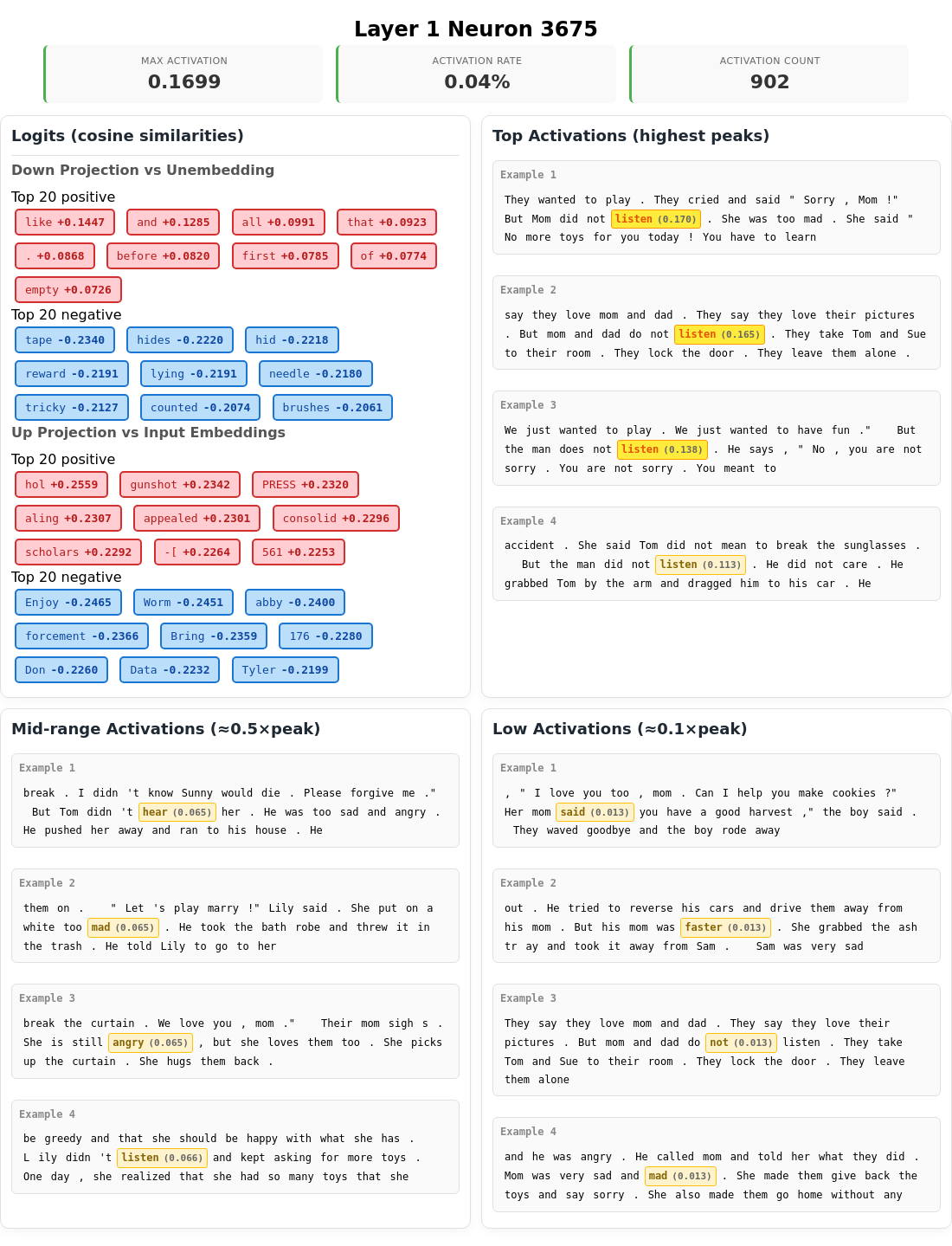}
\caption{\textbf{L1‑N3675} activates on the verb \emph{listen} in negated contexts (\emph{``did not listen''}, \emph{``does not listen''}), typically in scenes involving disobedience or conflict between characters. Negative logits include \emph{tape, hides, reward}.}
\end{figure*}

\clearpage
\begin{figure*}[h!]
\centering
\includegraphics[width=\textwidth]{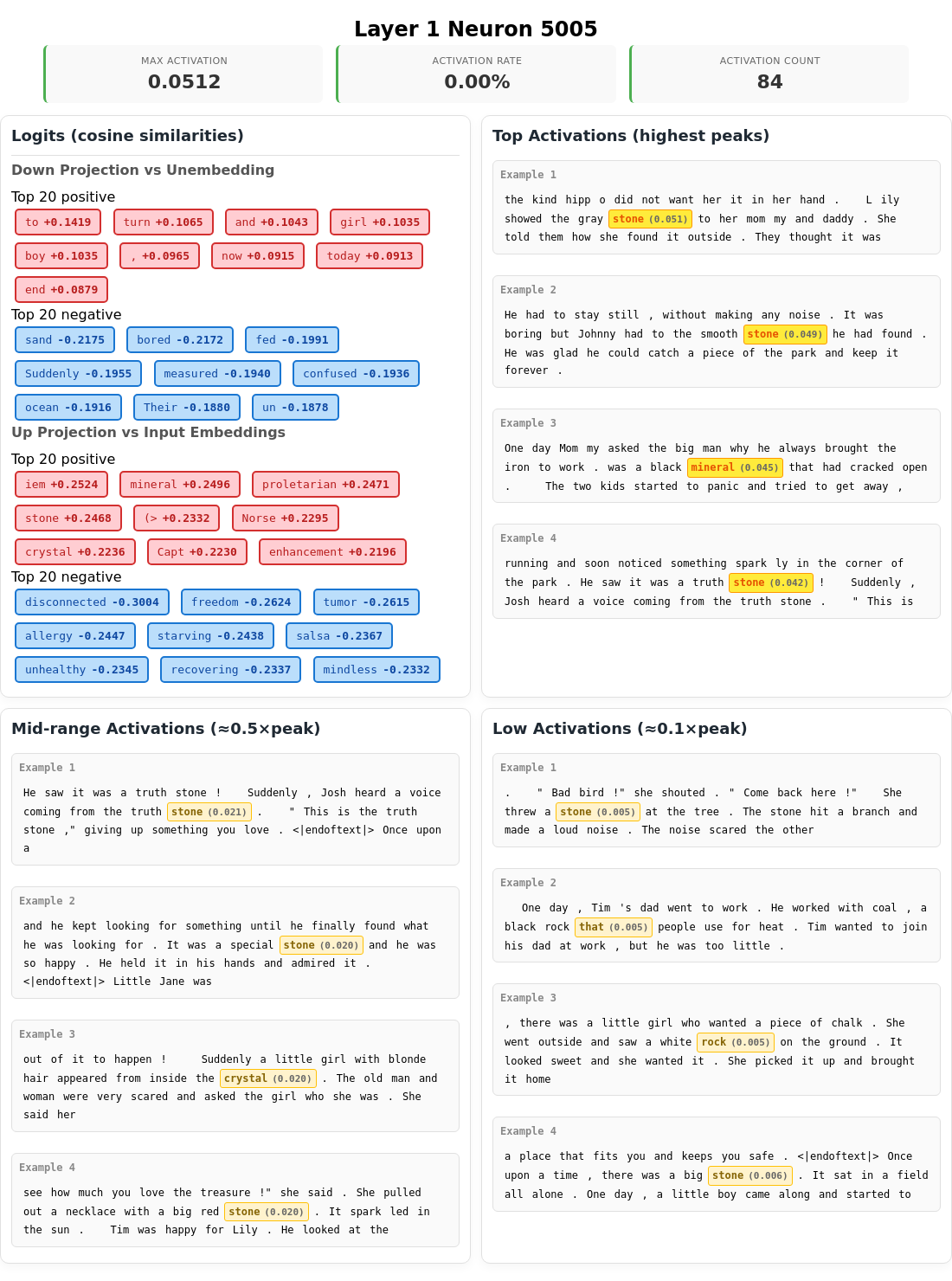}
\caption{\textbf{L1‑N5005} detects geological objects: \emph{stone, mineral, crystal}. Its top examples consistently mention finding or holding a stone.}
\end{figure*}

\clearpage
\begin{figure*}[h!]
\centering
\includegraphics[width=\textwidth]{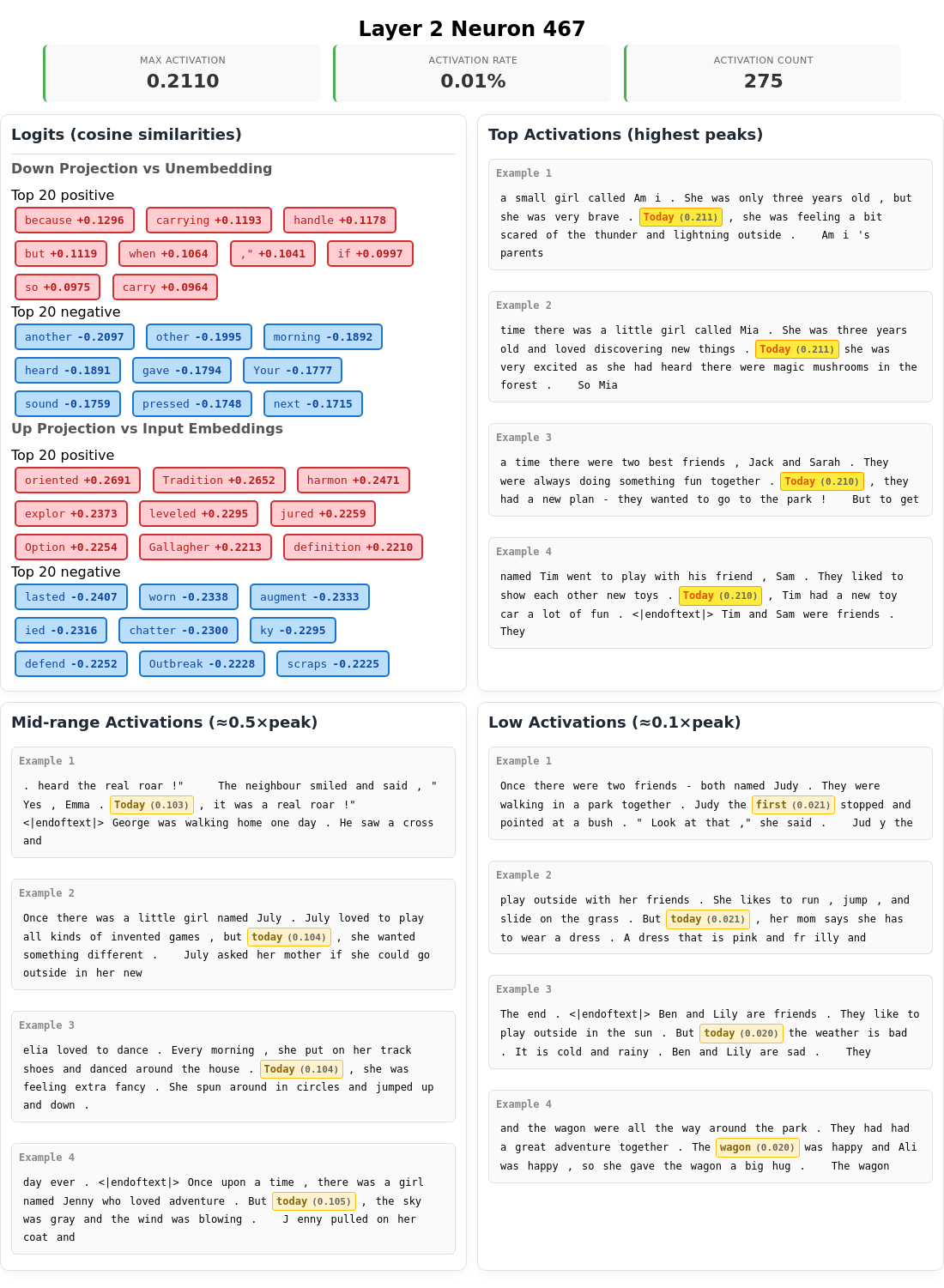}
\caption{\textbf{L2‑N467} marks the word \emph{Today} at the beginning of a narrative sentence, often followed by a comma, signalling a temporal shift in the story.}
\end{figure*}

\clearpage
\begin{figure*}[h!]
\centering
\includegraphics[width=\textwidth]{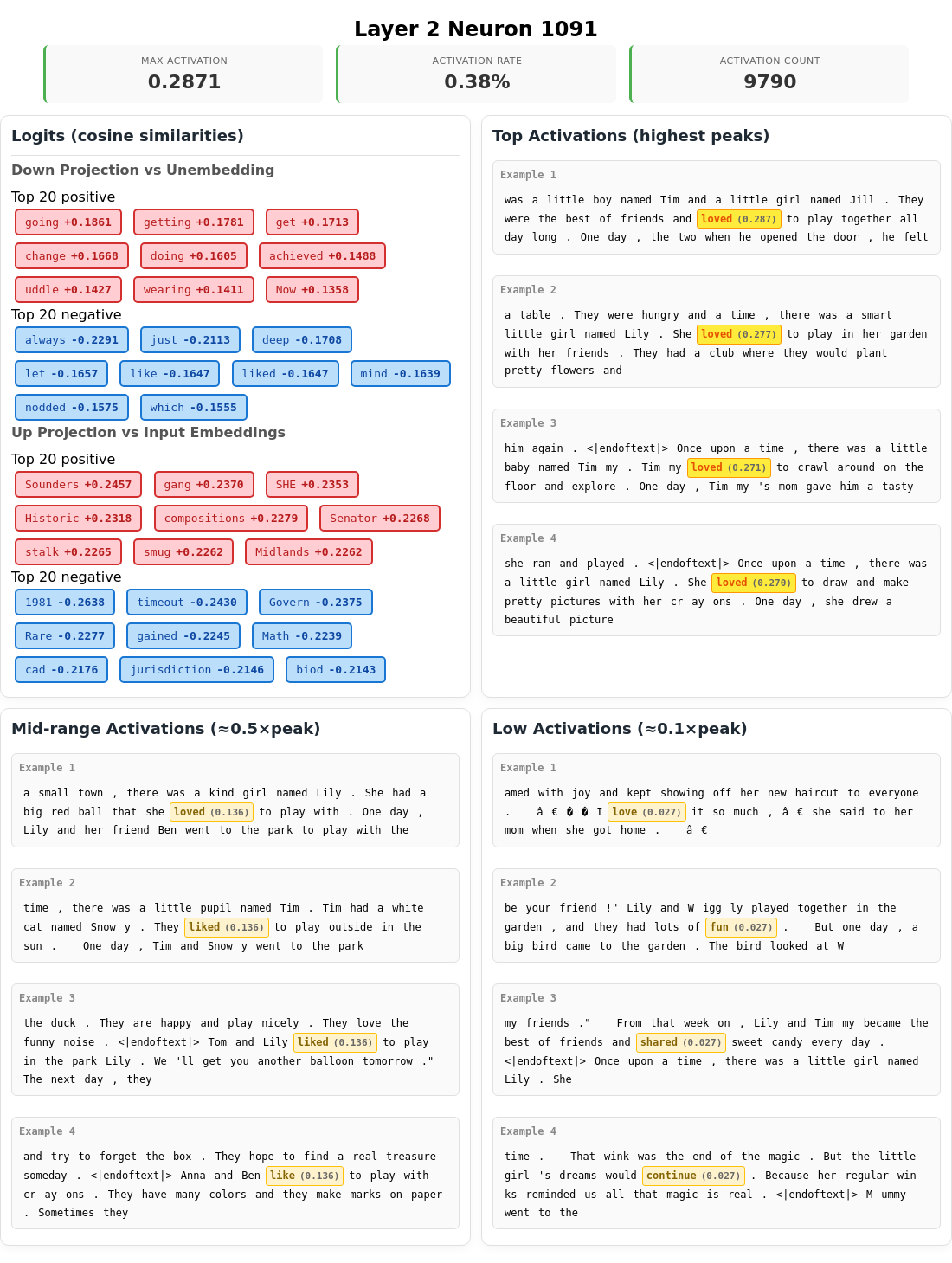}
\caption{\textbf{L2‑N1091} encodes a character's enjoyment of an activity: top phrases are \emph{``loved to play''}, \emph{``loved to crawl''}, \emph{``loved to draw''}.}
\end{figure*}

\clearpage
\begin{figure*}[h!]
\centering
\includegraphics[width=\textwidth]{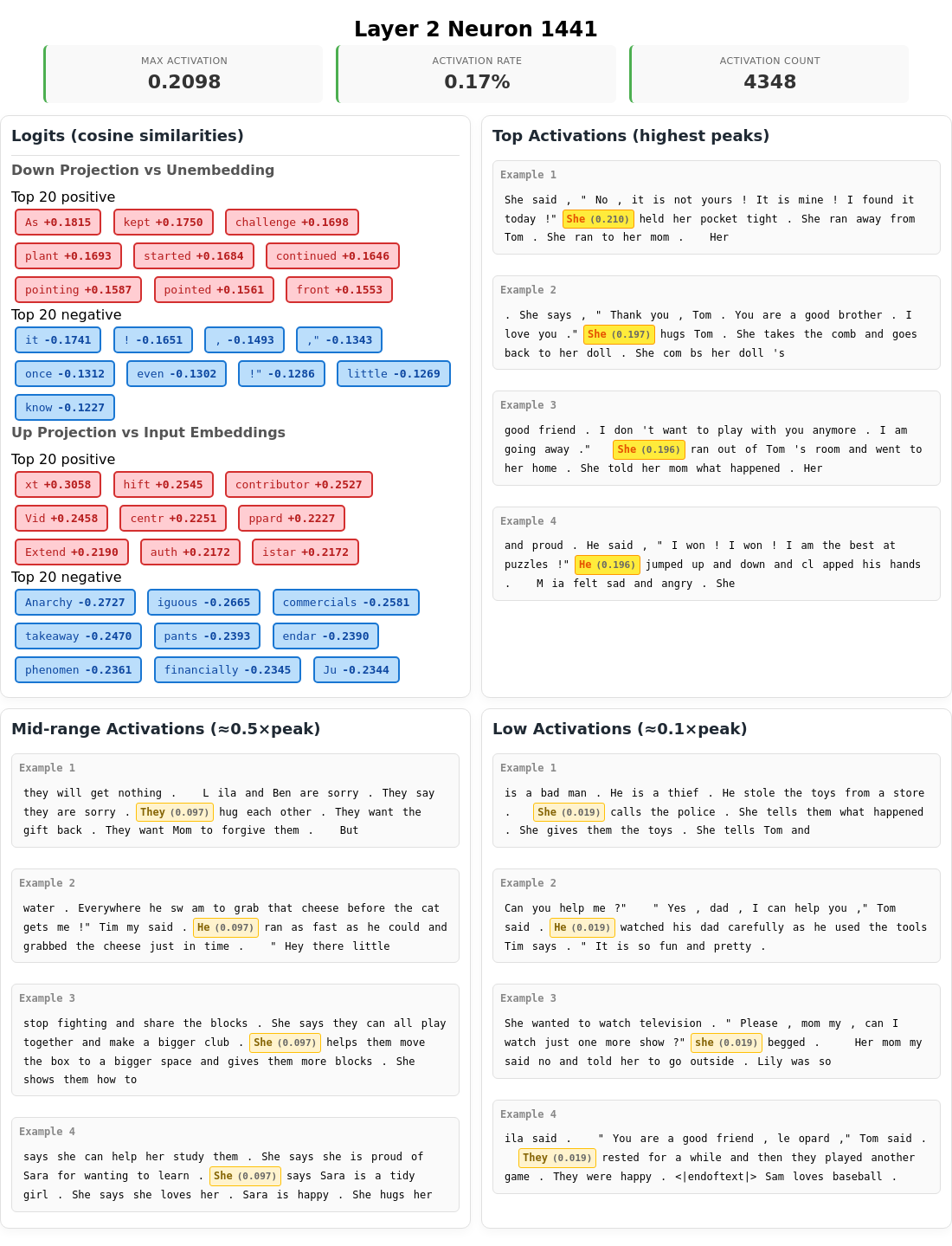}
\caption{\textbf{L2‑N1441} activates on sentence‑initial pronouns (\emph{She, He, They}) that introduce a new action or dialogue, e.g.\ \emph{``She held her pocket tight''}.}
\end{figure*}

\clearpage
\begin{figure*}[h!]
\centering
\includegraphics[width=\textwidth]{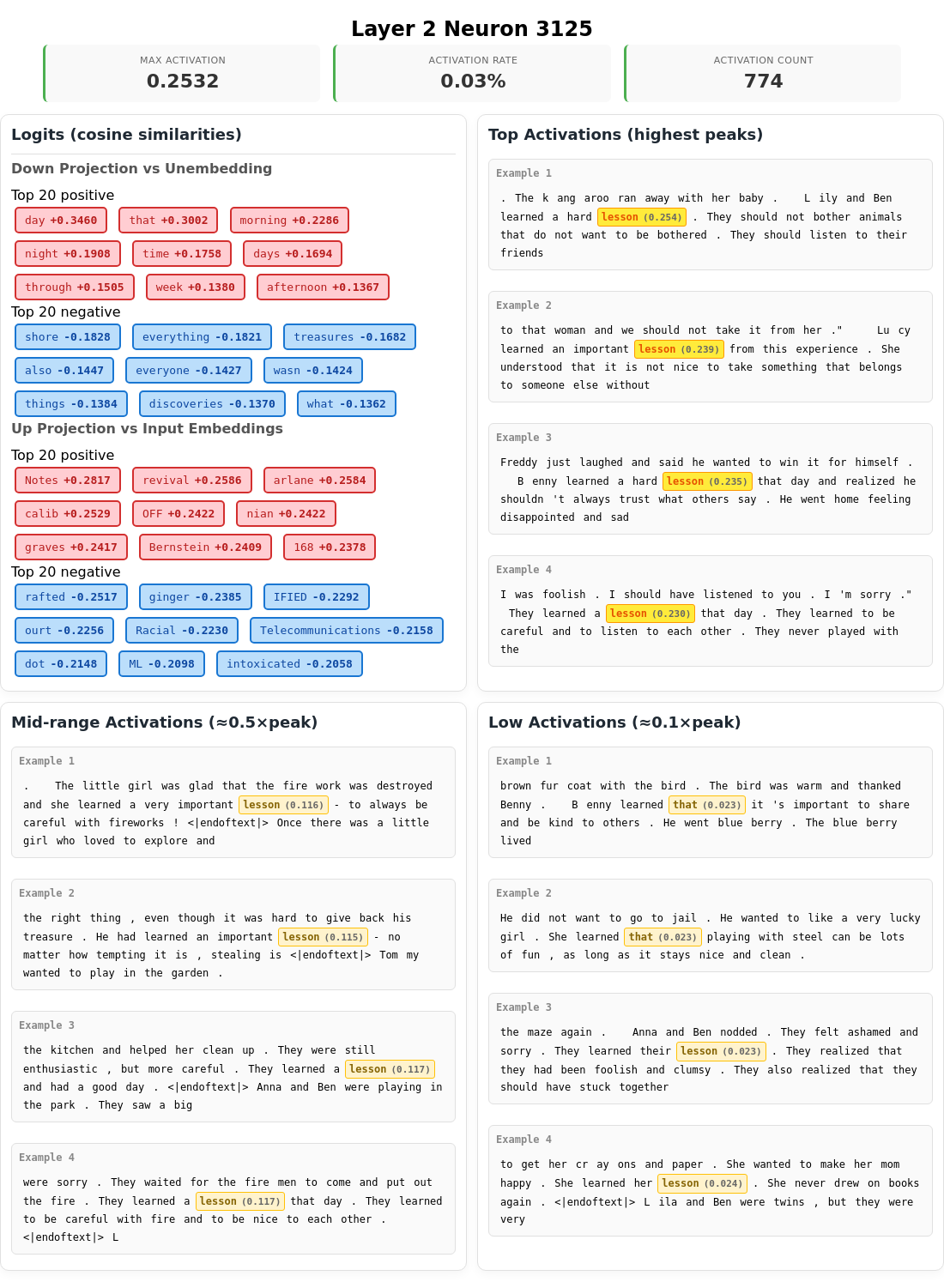}
\caption{\textbf{L2‑N3125} detects a moral or lesson learned: top activations contain \emph{``learned a lesson''}, \emph{``learned a hard lesson''}, often at the end of a story.}
\end{figure*}

\clearpage
\begin{figure*}[h!]
\centering
\includegraphics[width=\textwidth]{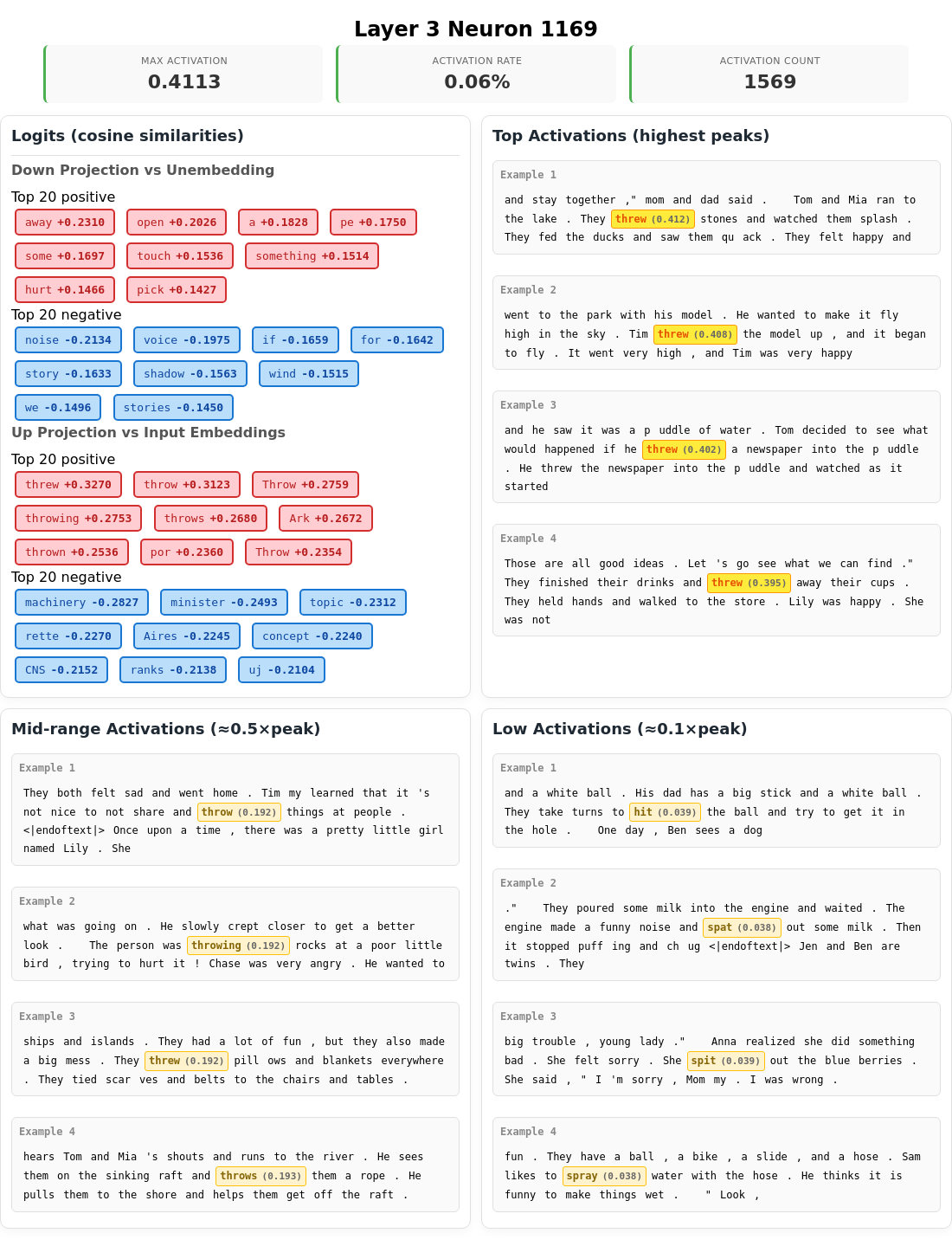}
\caption{\textbf{L3‑N1169} represents the action of throwing: top examples are \emph{``threw stones''}, \emph{``threw the model''}, \emph{``threw a newspaper''}.}
\end{figure*}

\clearpage
\begin{figure*}[h!]
\centering
\includegraphics[width=\textwidth]{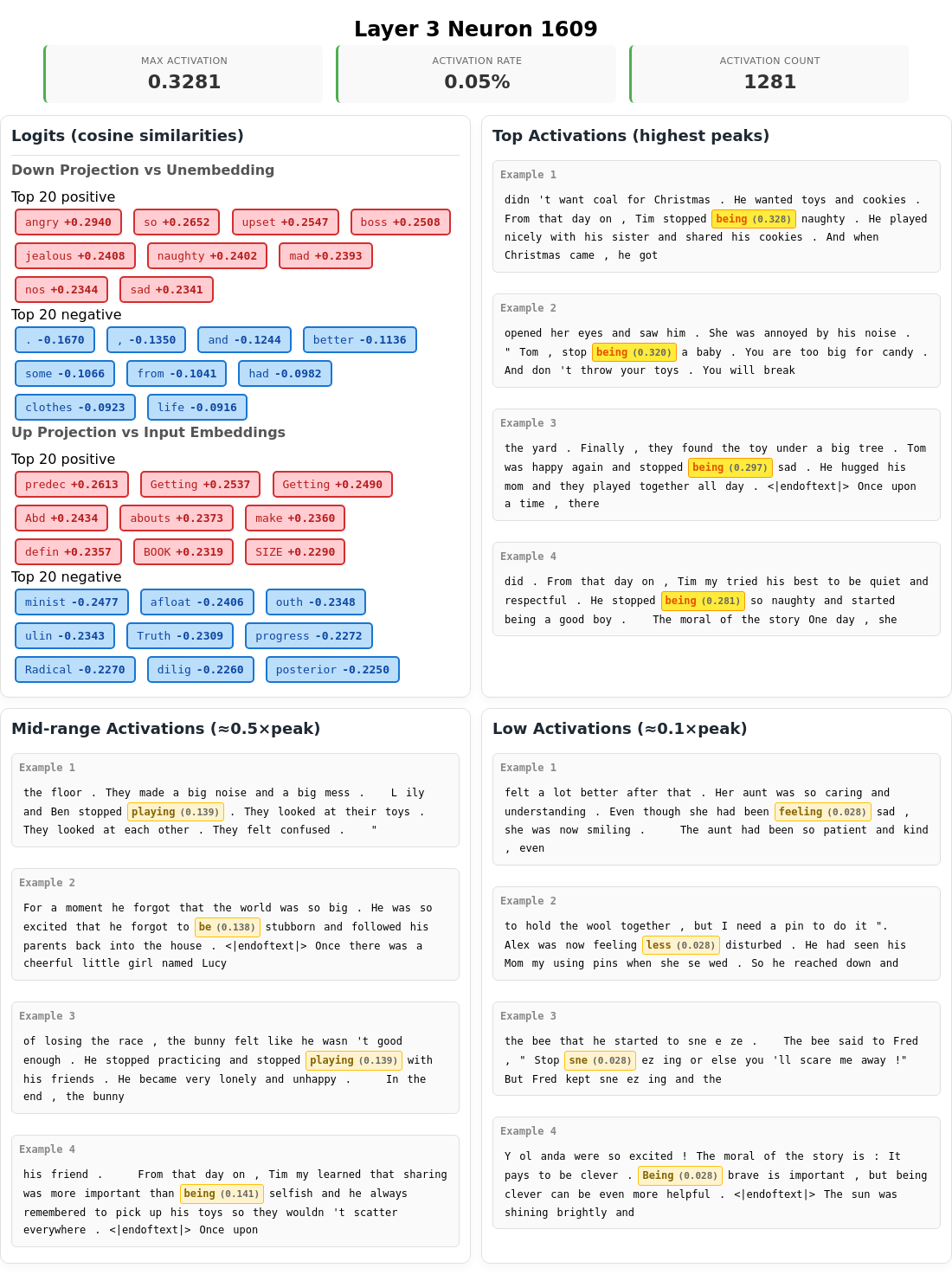}
\caption{\textbf{L3‑N1609} activates on the gerund \emph{being} at moments of emotional or behavioural shift in a character - particularly when a character is described as \emph{being naughty}, \emph{being stubborn}, or \emph{stopped being} a certain way. Its output weight vector promotes anger- and emotion-related tokens (\emph{angry, upset, jealous, naughty, mad}).}
\end{figure*}

\clearpage
\begin{figure*}[h!]
\centering
\includegraphics[width=\textwidth]{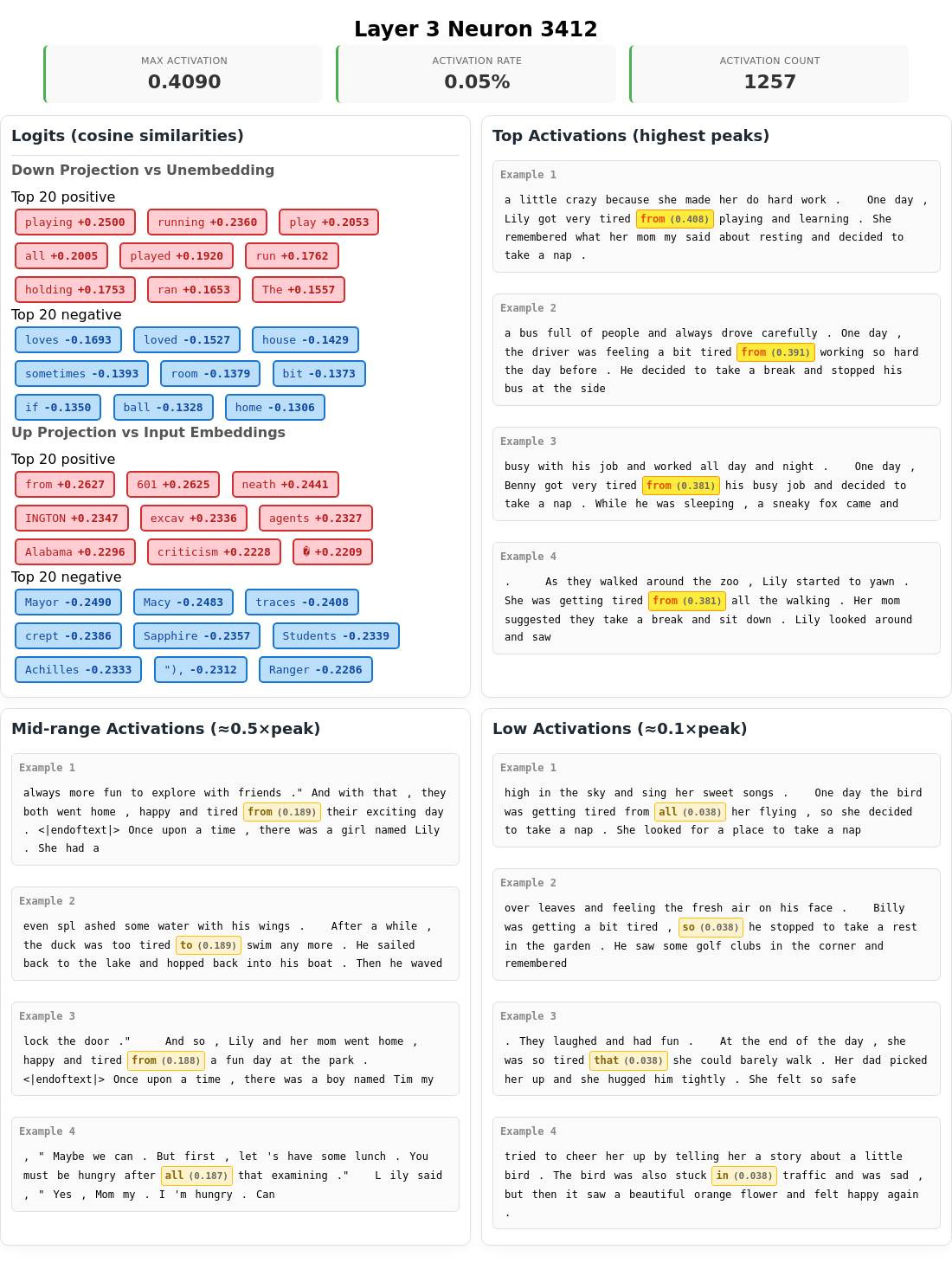}
\caption{\textbf{L3‑N3412} activates on the preposition \emph{from} in the construction \emph{``tired from [activity]''} (e.g.\ \emph{``tired from playing''}, \emph{``tired from working''}, \emph{``tired from all the walking''}), signalling physical exhaustion caused by a preceding effort.}
\end{figure*}

\clearpage
\begin{figure*}[h!]
\centering
\includegraphics[width=\textwidth]{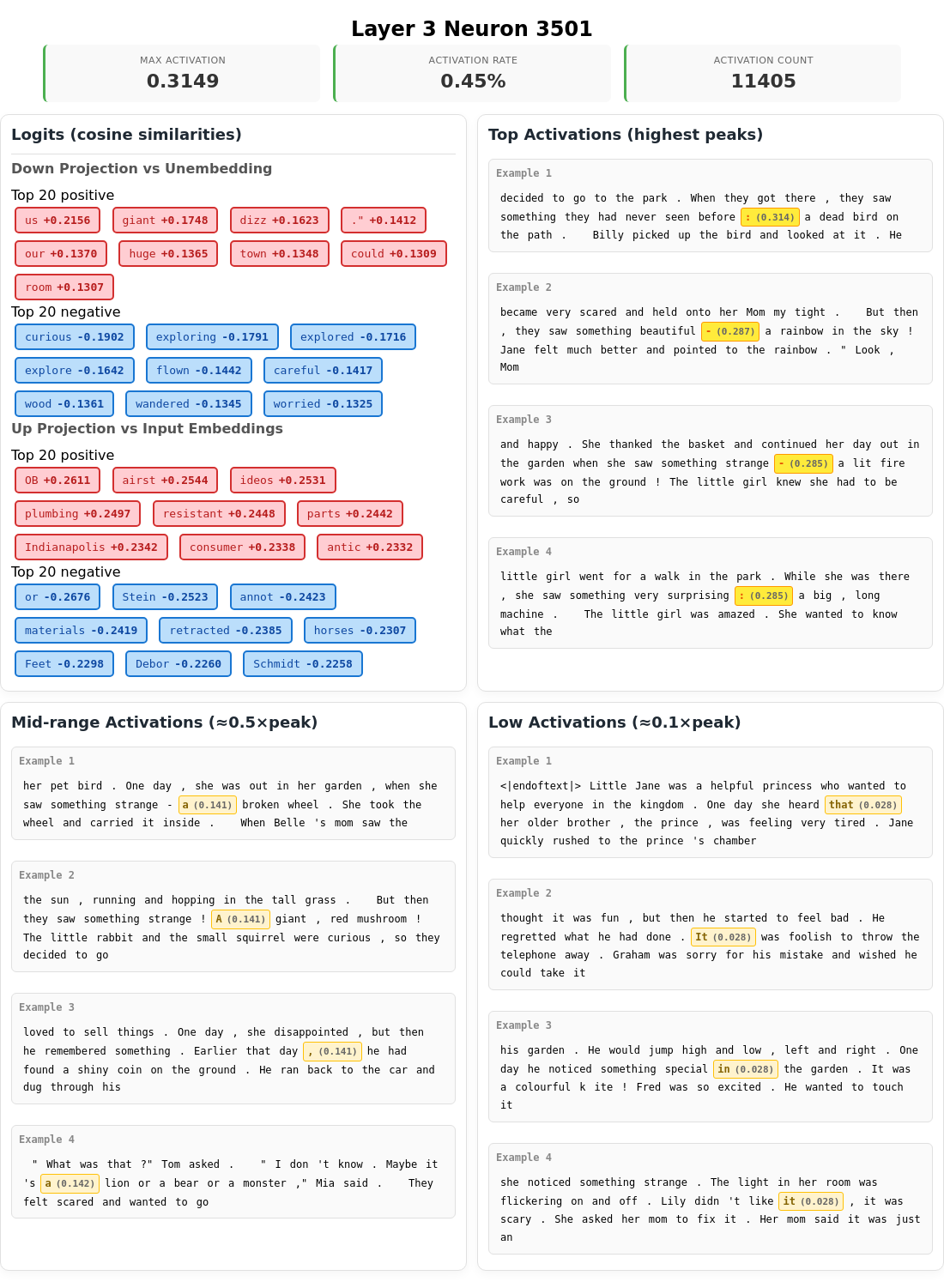}
\caption{\textbf{L3‑N3501} activates on the punctuation mark (colon or dash) that follows a phrase like \emph{``something they had never seen before''}, \emph{``something beautiful''}, or \emph{``something strange''} - the moment just before the surprising object is revealed or described.}
\end{figure*}

\clearpage
\begin{figure*}[h!]
\centering
\includegraphics[width=\textwidth]{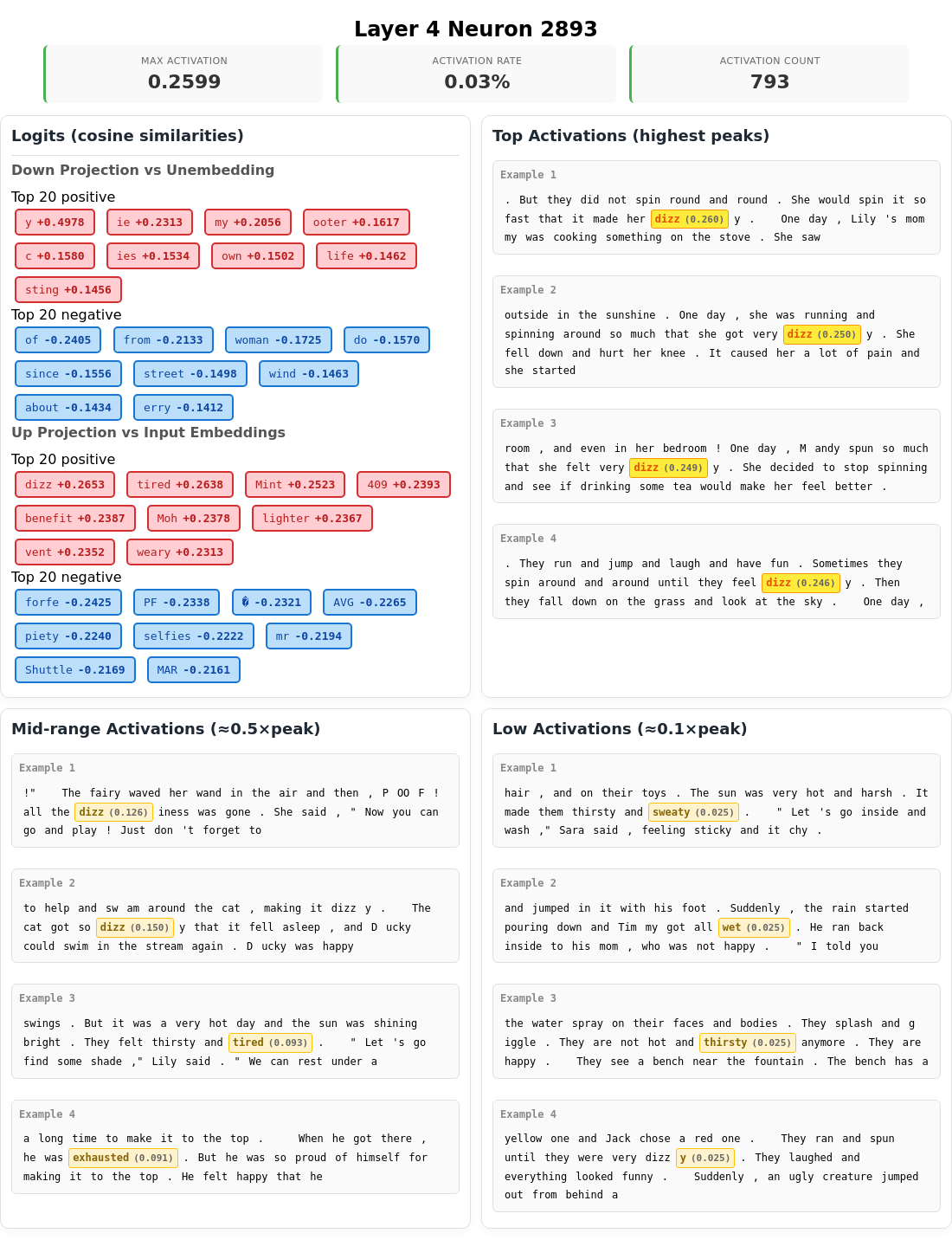}
\caption{\textbf{L4‑N2893} represents dizziness or spinning: top examples contain \emph{``dizzy''} (often split as \emph{``dizz y''}) from spinning too much.}
\end{figure*}

\clearpage
\begin{figure*}[h!]
\centering
\includegraphics[width=\textwidth]{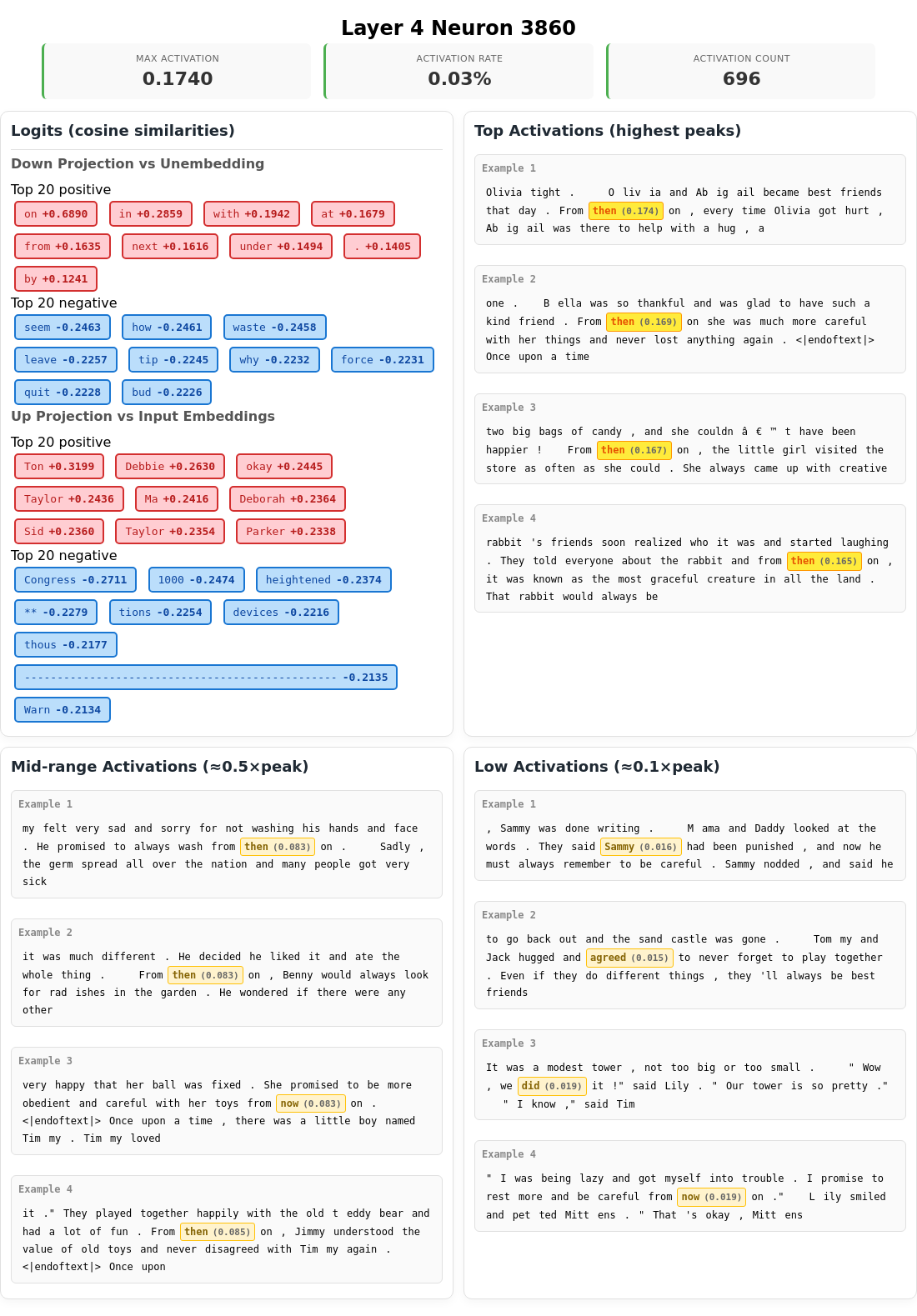}
\caption{\textbf{L4‑N3860} marks a temporal turning point: \emph{``From then on''}, \emph{``From now on''}, signalling that a character changes behaviour after an event.}
\end{figure*}

\clearpage
\begin{figure*}[h!]
\centering
\includegraphics[width=\textwidth]{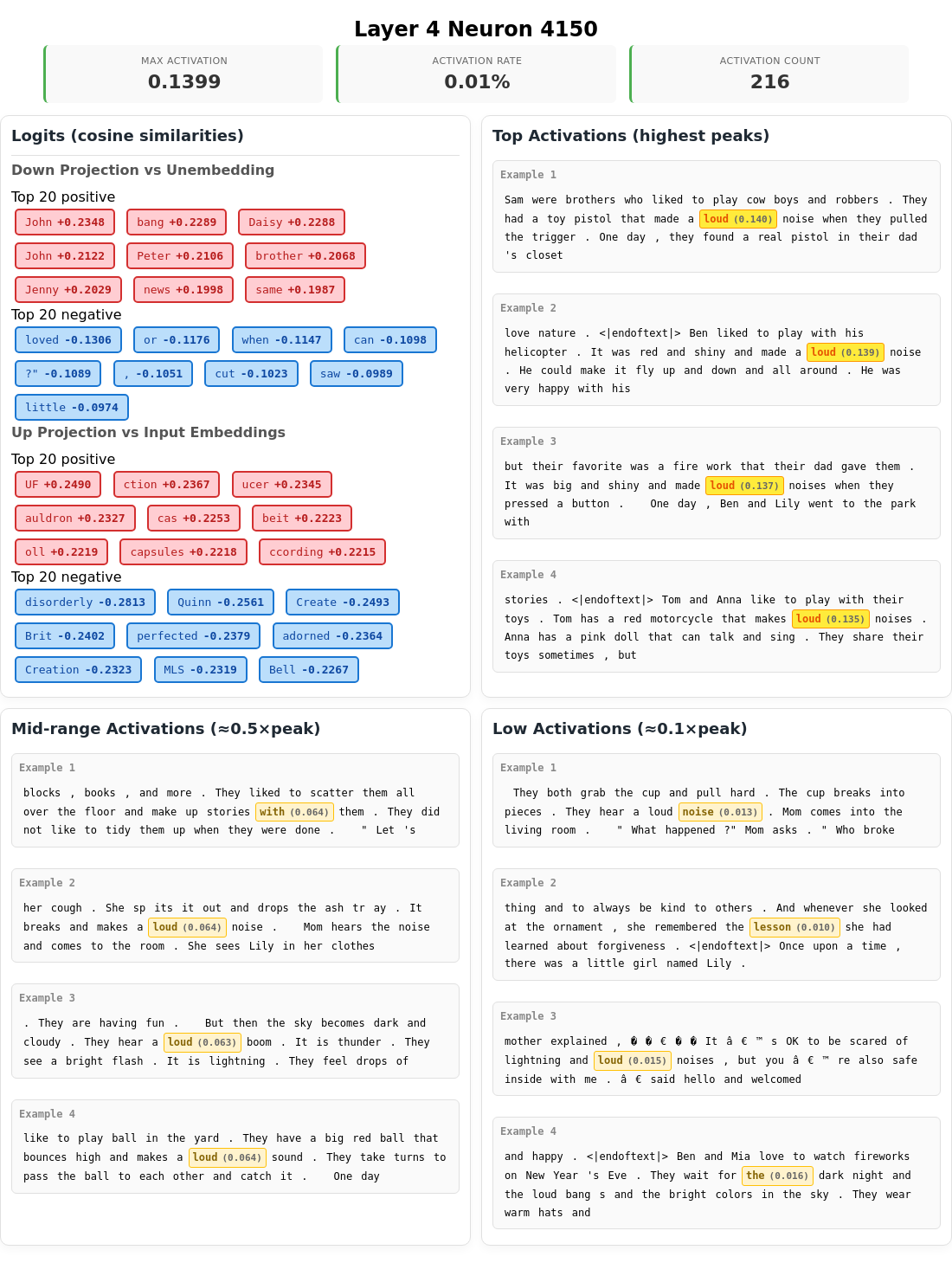}
\caption{\textbf{L4‑N4150} detects loud sounds: top activations are \emph{``loud noise''}, \emph{``loud bang''}, \emph{``loud noises''}, often from fireworks or broken objects.}
\end{figure*}

\clearpage
\begin{figure*}[h!]
\centering
\includegraphics[width=\textwidth]{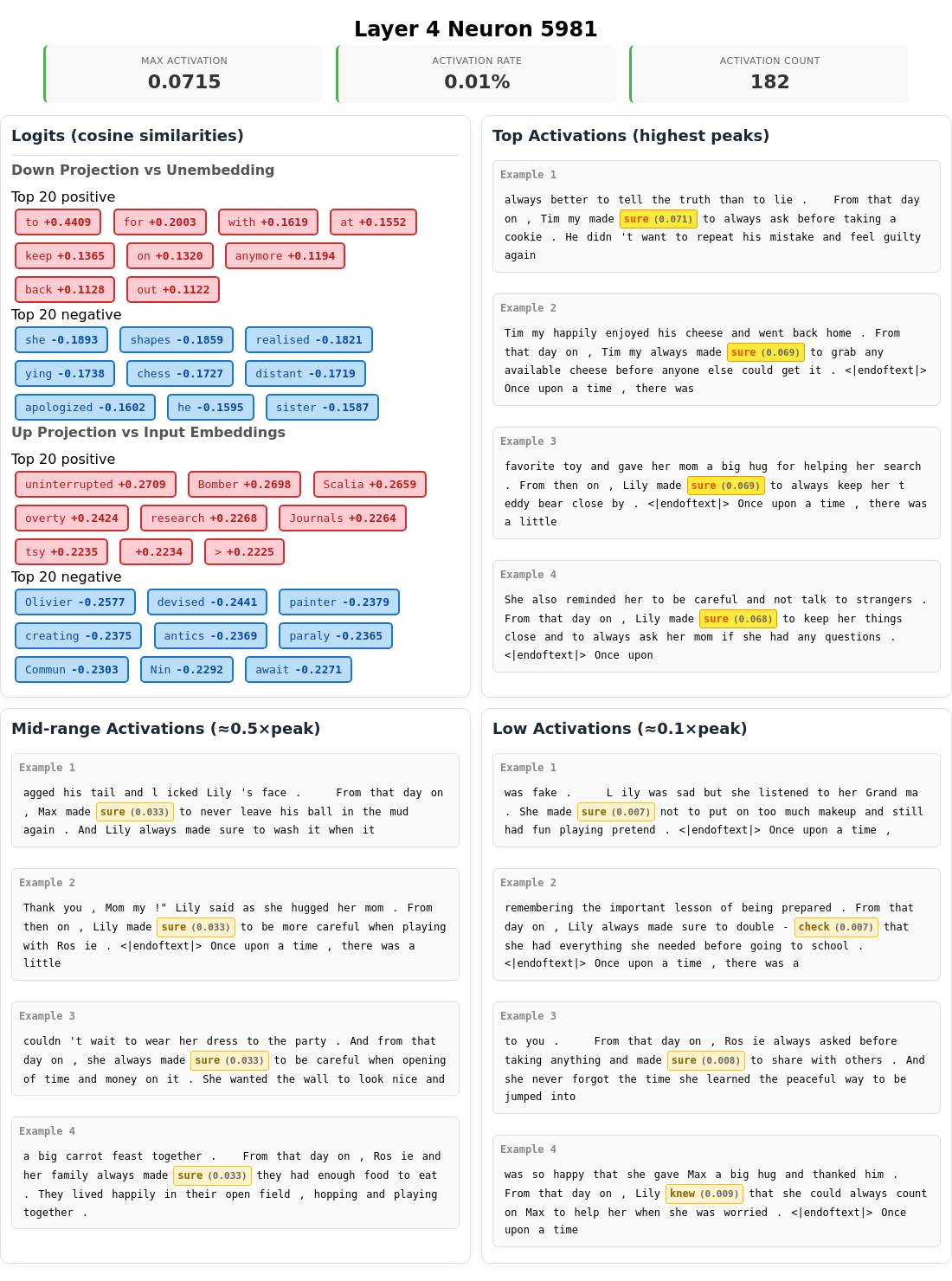}
\caption{\textbf{L4‑N5981} encodes a commitment or resolution: \emph{``From that day on, [character] made sure to always''} followed by a cautious behaviour.}
\end{figure*}

\clearpage
\begin{figure*}[h!]
\centering
\includegraphics[width=\textwidth]{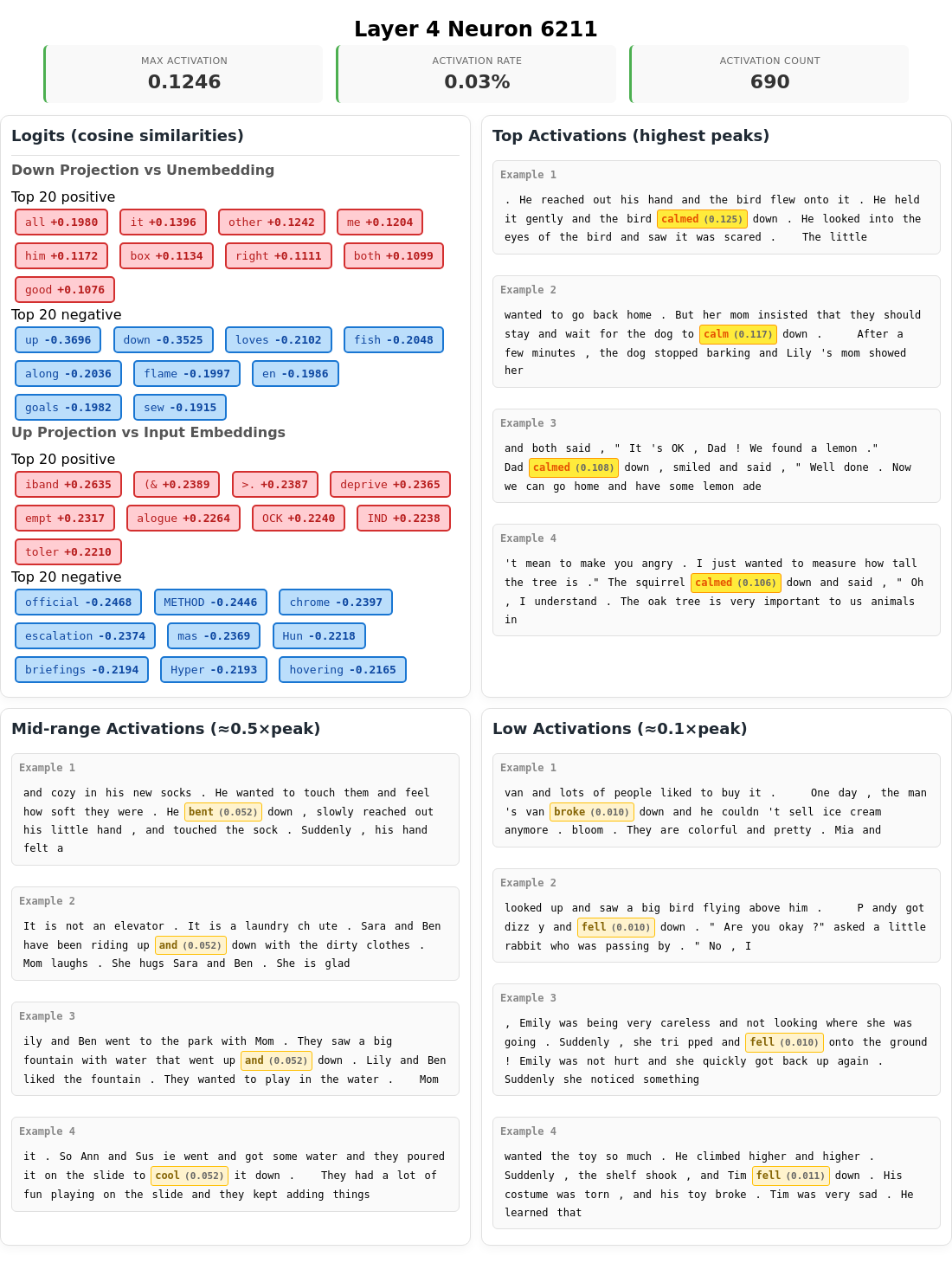}
\caption{\textbf{L4‑N6211} activates on verbs describing a downward or de-escalating motion (\emph{calmed, bent, fell, broke}) in the construction \emph{``[verb] down''}, particularly when a character or situation settles or physically descends. Negative logits include \emph{up, down} (as standalone tokens) and \emph{loves}.}
\end{figure*}

\end{document}